\documentclass{article}
\usepackage{iclr2027_conference,times}
\iclrfinalcopy 
\usepackage[T1]{fontenc}
\usepackage{amsmath,amssymb,mathtools,amsthm}
\theoremstyle{remark}
\newtheorem{remark}{Remark}
\usepackage{booktabs,array,tabularx,makecell}
\usepackage{graphicx,adjustbox,xcolor}
\usepackage{colortbl}
\definecolor{successcolumn}{HTML}{B6FF59}
\definecolor{failurecolumn}{HTML}{FFB0B0}
\usepackage{tikz}
\usetikzlibrary{arrows.meta,positioning,calc}

\usepackage{bm}
\usetikzlibrary{arrows.meta,positioning}
\usepackage{microtype,hyperref,url}
\usepackage{enumitem,float,placeins,needspace,flafter}
\definecolor{linkblue}{HTML}{5B9BD5}
\hypersetup{colorlinks=true,linkcolor=linkblue,citecolor=linkblue,urlcolor=linkblue,hypertexnames=false,
 pdftitle={Residual-Stream Burden Shapes Representation Learning in Diffusion Transformers},
 pdfauthor={Tongtong Liang, Siqi Kou, Ziqiao Xi, Esha Singh, Kun Zhou, Zhijie Deng, Alexander Cloninger, Yu-Xiang Wang, Rahul Parhi}}

\newsavebox{\burdenboxcontent}
\newenvironment{concepthighlight}{%
  \par\smallskip\begingroup\setlength{\fboxsep}{5pt}%
  \begin{lrbox}{\burdenboxcontent}%
  \begin{minipage}{\dimexpr\linewidth-2\fboxsep\relax}%
}{%
  \end{minipage}\end{lrbox}%
  \noindent\colorbox{black!5}{\usebox{\burdenboxcontent}}\par\endgroup\smallskip
}
\newcommand{\er}{\operatorname{ER}}

\newcommand{\Cov}{\operatorname{Cov}}
\renewcommand{\vec}[1]{{\bm{#1}}}
\newcommand{\Hpre}{\bm{\mathcal H}_{\mathrm{pre}}}
\newcommand{\Hpost}{\bm{\mathcal H}_{\mathrm{post}}}
\newcommand{\Hmix}{\bm{\mathcal H}_{\mathrm{mix}}}

\def\mA{{\bm{A}}}

\def\mG{{\bm{G}}}
\def\mH{{\bm{H}}}
\def\mI{{\bm{I}}}
\def\mJ{{\bm{J}}}

\def\mP{{\bm{P}}}

\def\mU{{\bm{U}}}
\def\mV{{\bm{V}}}
\def\mW{{\bm{W}}}

\newcommand{\dd}{\,\mathrm d}

\newcommand{\E}{\mathbb{E}}

\setlist[itemize]{leftmargin=1.4em,itemsep=3pt,topsep=3pt}
\setlist[enumerate]{leftmargin=1.7em,itemsep=4pt,topsep=3pt}
\title{Residual-Stream Burden Shapes Representation Learning in Diffusion Transformers}
\author{%
\begin{minipage}{\dimexpr\textwidth-2\tabcolsep\relax}
\raggedright\normalfont
\textbf{Tongtong Liang}\textsuperscript{1}\quad
\textbf{Siqi Kou}\textsuperscript{2}\quad
\textbf{Ziqiao Xi}\textsuperscript{3}\quad
\textbf{Esha Singh}\textsuperscript{1}\quad
\textbf{Kun Zhou}\textsuperscript{3}\\[3pt]
\textbf{Zhijie Deng}\textsuperscript{2}\quad
\textbf{Alexander Cloninger}\textsuperscript{1}\quad
\textbf{Yu-Xiang Wang}\textsuperscript{1}\quad
\textbf{Rahul Parhi}\textsuperscript{1}\\[7pt]
\textsuperscript{1}UC San Diego\quad
\textsuperscript{2}Shanghai Jiao Tong University\quad
\textsuperscript{3}Aether AI
\end{minipage}}
\begin{document}
\raggedbottom
\maketitle
\fancyhead[L]{Preprint}

\begin{abstract}
In diffusion-based generation, a neural network can be trained to predict the clean data, the noise, or the velocity from a noisy input.
These prediction targets are interconvertible and describe the same generative process, yet plain Diffusion Transformers operating on large pixel patches succeed with clean prediction and fail with noise or velocity prediction.
We argue that this asymmetry arises because noisy targets require the residual stream to preserve noise-dependent input variation through depth for the final readout, forcing subsequent layers to compute on noisy representations.
A spectrally concentrated clean target imposes a lighter demand, leaving greater freedom to organize hidden representations for subsequent computation.
We call this preservation requirement \emph{residual-stream burden} and show how it shapes representation learning in Diffusion Transformers.
Controlled experiments indicate that the exploitable structure is spectral concentration in patch space and that the bandwidth of the persistent residual state is a key resource for noisy prediction.
We further show that this account is consistent with recent decoupled pixel-space architectures, whose diverse designs all reduce the residual-stream burden on the main pathway.
To examine this understanding from a complementary direction, we expand and reorganize the residual-stream bandwidth directly, introducing Spatially Indexed Hyper-Connections (SiHC) that reach FID 1.71 on ImageNet $256^2$.
Together, these results identify residual-stream burden as a mechanism through which prediction targets and architecture jointly shape representation learning in Diffusion Transformers.
\end{abstract}

\begin{center}
\small
\textbf{Codebase:} \url{https://github.com/tongtongliang/residual-stream-burden}\\[2pt]
\textbf{Project page:} \url{https://tongtongliang.github.io/residual-stream-burden/}
\end{center}

\section{Introduction}\label{sec:intro}
Denoising generative models learn a data distribution by training a neural network to separate clean data from its corrupted version.
Given a noisy input, one can ask the network to predict the noise ($\vec{\epsilon}$-prediction) \citep{ho2020ddpm}, the velocity ($\vec{v}$-prediction, a linear combination of clean data and noise) \citep{salimans2022progressive,lipman2023flow,liu2023rectified}, or the clean data itself ($\vec{x}$-prediction) \citep{li2026back}.
These parameterizations describe the same generative process and can be converted into one another at inference time, so one might expect them to present comparable learning problems.
However, when plain Diffusion Transformers (DiTs) \citep{peebles2023dit} operate on large image patches, the choice becomes decisive: with $16\times 16$ patches, $\vec{x}$-prediction succeeds while the other two fail catastrophically \citep{li2026back}.

A recent explanation attributes the advantage of $\vec{x}$-prediction to the \emph{low-dimensional manifold} of natural images, arguing that this structure allows clean prediction to use less \emph{model capacity} while noisy targets force the network to reproduce high-dimensional noise \citep{li2026back}.
But this account leaves two key concepts imprecise: what kind of low-dimensionality does the network exploit, what exactly is the capacity consumed by noisy prediction, and how do the two interact during training?
We answer these questions and show how the resulting understanding informs architecture design for diffusion models.

\textbf{Our contributions.}
We propose that the advantage of $\vec{x}$-prediction arises because noisy targets require a plain DiT's hidden states to retain noise-dependent input variation through the residual stream for the final readout, so that subsequent layers must compute on noisy representations.
As patch size grows, more corruption-dependent pixel variation is packed behind each token, increasing the preservation pressure on its residual state.
By contrast, a clean target concentrated in fewer directions reduces the demand on \emph{information-carrying capacity} for the final readout.

More generally, the output target determines what information must be preserved through depth, and this preservation requirement shapes the representations learned along the residual stream.
We call this requirement \emph{residual-stream burden}.
Accordingly, the relevant low-dimensionality should appear in the patch space through which a DiT sees the image, and the relevant capacity should be the bandwidth of the persistent state that carries information across depth. Specifically, we find that

\begin{itemize}[leftmargin=*]
\item \emph{The exploitable low-dimensionality is spectral concentration in patch space.}
We show that the exploitable structure is the spectral concentration of clean-patch variance: an invertible whitening of the patch space that preserves manifold structure but removes this concentration destroys the success of $\vec{x}$-prediction (Section \ref{sec:patch-geometry}).
We further show that this concentration shapes representation learning differently under $\vec{x}$- and $\vec{v}$-prediction (Section \ref{sec:target-burden}).

\item \emph{The relevant capacity is residual-stream bandwidth.}
Using Hyper-Connections (HC) \citep{zhu2025hyperconnections} to expand the persistent residual state $4\times$ while keeping computation nearly unchanged, we rescue $\vec{v}$-prediction from complete failure, indicating that residual-stream bandwidth is a key resource for accommodating the burden of noisy prediction (Section \ref{sec:hc-probe}).
\end{itemize}

This understanding also accounts for the success of several recent pixel-space architectures that predict velocity through decoupled designs, such as DeCo, PixelDiT, and DiP \citep{ma2026deco,yu2026pixeldit,chen2026dip}, whose separate pixel paths absorb fine-scale noise prediction and relieve the main backbone's burden.
We verify that their backbone streams develop the same concentrated structure and representation-learning behavior as $\vec{x}$-prediction in a plain DiT (Section~\ref{sec:architecture-reconciliation}).

To further examine this understanding, we explore a complementary direction: instead of relieving burden through a separate decoder or switching to $\vec{x}$-prediction, we reorganize the residual-stream bandwidth itself.
We introduce \emph{Spatially Indexed Hyper-Connections} (SiHC), which expand the persistent bandwidth and distribute the prediction burden across local subpatch states.
Under direct $\vec{v}$-prediction without a specialized decoder or cross-attention, SiHC-XL with REPA \citep{yu2025repa} reaches FID 1.71 on ImageNet $256^2$, competitive with recent pixel-space methods.

\section{Related Work}\label{sec:related-work}
\textbf{Decoupled designs in diffusion models.}
Decoupled designs separate semantic or low-frequency modeling from the fine-scale reconstruction required by the diffusion target, an idea made explicit by DDT through a semantic encoder and a specialized velocity decoder \citep{wang2026ddt}. The design has since been adopted in pretrained representation spaces such as RAE \citep{zheng2026rae} and in recent pixel-space models (Table~\ref{tab:pixel-paths}). Prediction reparameterization can be viewed as a fixed, analytic form of decoupling, in which $\vec x$-prediction supplies the noisy input through an analytic skip \citep{li2026back,chen2026asymflow}. Our analysis shows that these seemingly different designs share the effect of reducing the noise-dependent burden on the main DiT, thereby easing the friction between velocity modeling and representation learning. SiHC achieves this separation without a separate pixel decoder or cross-attention path, expanding and organizing residual-stream bandwidth so that local states carry the burden outside the shared workspace.

\begin{table}[t]
\centering\setlength{\tabcolsep}{6pt}\renewcommand{\arraystretch}{1}
\caption{\textbf{A taxonomy of pixel-space diffusion architectures.} Selected works. More details are in Appendix \ref{app:related}.}
\label{tab:pixel-paths}\label{fig:rw-taxonomy}
\begin{tabularx}{\linewidth}{@{}>{\hsize=1\hsize\linewidth=\hsize\raggedright\arraybackslash}X>{\hsize=1.00\hsize\linewidth=\hsize\raggedright\arraybackslash}X>{\hsize=1.00\hsize\linewidth=\hsize\raggedright\arraybackslash}X@{}}
\toprule
\centering\makecell{\small\textbf{Hierarchical design}} & \centering\makecell{\small\textbf{Decoupling via long skip}} & \centering\arraybackslash\makecell{\small\textbf{Decoupling via cross-attention}} \\
\midrule
\includegraphics[width=\linewidth,height=90pt,keepaspectratio]{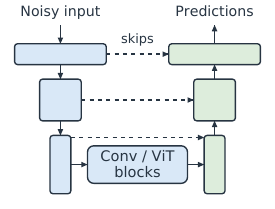} &
\includegraphics[width=\linewidth,height=90pt,keepaspectratio]{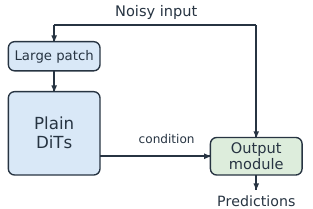} &
\includegraphics[width=\linewidth,height=90pt,keepaspectratio]{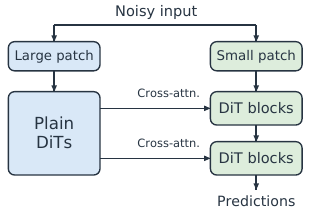} \\
\midrule
\tiny ADM \citep{dhariwal2021adm}, 
SiD / SiD2 \citep{hoogeboom2023simple,hoogeboom2025sid2}, PixelFlow \citep{chen2025pixelflow}&\tiny
PixNerd \citep{wang2026pixnerd}, 
DiP \citep{chen2026dip}, 
DeCo \citep{ma2026deco}, 
PixelDiT \citep{yu2026pixeldit}
&\tiny
RIN \citep{jabri2023rin}, 
HyperDiT \citep{he2026hyperdit}, 
DuSPiT \citep{bai2026duspit} \\
\bottomrule
\end{tabularx}
\end{table}

\textbf{Hyper-Connections and virtual width.}
Hyper-Connections (HC) replace the single residual stream with multiple persistent streams and learned cross-depth connections \citep{zhu2025hyperconnections}, allowing representational width to grow independently of computational width, a degree of freedom termed \emph{virtual width} \citep{seed2025virtualwidth}. Follow-up work has made expanded residual streams stable and scalable, and HC variants now appear in large language models \citep{zhu2025fracconnections,xie2025mhc,zhang2026xhc,zai2026glm53card,tencent2026hy4}. Their role in generative modeling remains much less understood, and the original HC work reports only a modest improvement on latent DiT-XL/2 (FID $2.36\!\rightarrow\!2.18$) \citep{zhu2025hyperconnections}. More recently, Diffusion-Adaptive Routing studies how cross-layer aggregation affects optimization in diffusion Transformers and improves training through timestep-aware access to earlier layer outputs, routing information along depth rather than widening the persistent state \citep{xu2026routing}. Our work provides a demand-side view of residual-stream bandwidth from the perspective of residual-stream burden and shows that accommodating this burden with virtual width yields particularly large gains in high-dimensional flow matching.

\textbf{Patch geometry and neural-network training.}
Natural-image patch geometry has long served as a prior in image processing and vision \citep{dabov2007bm3d,peyre2009manifold,yu2012piecewise,coates2011singlelayer,thiry2021patches,brutzkus2022patchfeatures}, rather than as a factor in how neural networks train. From the implicit-bias perspective, \citet{liang2026sparseconnectivity} characterize how \emph{input-side} patch geometry affects gradient descent in two-layer patch-based ReLU networks. Inspired by this viewpoint, we study the complementary \emph{output-side} patch geometry in modern diffusion Transformers and trace how it shapes hidden representations across depth.

\textbf{Representations for generation.}
Diffusion models learn useful visual features \citep{xiang2023ddae,tang2023dift,luo2023diffusionhyperfeatures,chen2025deconstruct,stracke2025cleandift}, though weaker than those of pretrained visual encoders \citep{yu2025repa}, and stronger intermediate representations are closely associated with faster training and better generation \citep{yu2025repa,leng2025repae,wu2025reg,singh2026irepa}. Our work suggests that one source of this representation-learning difficulty is \emph{residual-stream burden}, which can create friction between noisy-target prediction and representation learning. The RAE family instead performs diffusion directly in pretrained representation spaces and converges much faster \citep{zheng2026rae,singh2026raev2}. Section~\ref{sec:discussion} discusses how residual-stream burden bears on this advantage, and Appendix~\ref{sec:rw} gives further background.

\section{Background and Problem Setup}\label{sec:preliminaries}

\noindent\textbf{Flow matching.} We study the ODE formulation of flow-based models, where a neural network parameterizes a time-dependent velocity field \citep{lipman2023flow,liu2023rectified}. Let $\vec x\sim p_{\mathrm{data}}$ and $\vec\epsilon\sim\mathcal N(0,\mI)$ be independent. We use the linear interpolation
\begin{equation}\vec x_t=t\vec x+(1-t)\vec\epsilon,\qquad \vec v=\dd\vec x_t/\dd t=\vec x-\vec\epsilon,\qquad t\in[0,1].\label{eq:linear-flow}\end{equation}
Thus $t=0$ is noise and $t=1$ is clean data. The network is trained with $\mathcal L_{\mathrm{FM}}=\E_{t,\vec x,\vec\epsilon}\|\vec v_\theta(\vec x_t,t)-\vec v\|_2^2$, using the training time distribution on $(0,1)$. For $t<1$, the target also satisfies $\vec v=(\vec x-\vec x_t)/(1-t)$. Under $\vec v$-prediction the network outputs $\vec v_\theta$ directly, and under $\vec x$-prediction it outputs the clean image $\vec x_\theta$, which is converted as $\vec v_\theta=(\vec x_\theta-\vec x_t)/(1-t)$ \citep{li2026back}. Sampling solves $\dd\vec x_t/\dd t=\vec v_\theta(\vec x_t,t)$ from Gaussian noise at $t=0$ to data at $t=1$, by default with a 50-step Heun solver following \citet{li2026back}. Class conditioning is omitted from the notation.

\noindent\textbf{Residual stream.} A residual network passes a running representation through depth, updating it recursively as $\vec h_{\ell+1}=\vec h_\ell+\mathcal F_\ell(\vec h_\ell,t)$ \citep{he2016resnet}. Starting from the input embedding $\vec h_0$, each attention or MLP sublayer reads the current representation and adds its output to it. This sequence of representations forms the \emph{residual stream}, in which information carried forward by the identity path also supplies the input to subsequent computation. In a plain Transformer, the width-$C$ stream itself serves as each layer's computational \emph{workspace}.

We now introduce the central notion of this paper.
\begin{concepthighlight}
\textbf{Residual-stream burden} is the requirement to preserve prediction-relevant input information through depth for the final readout.
\end{concepthighlight}

\textbf{A toy experiment to visualize the residual-stream burden.} Following \citet[Section~3.3]{li2026back}, the synthetic ground-truth data are a planar Swiss roll embedded in $\mathbb{R}^{512}$. The orthogonal embedding matrix is unknown to the model. We train toy flow-matching models to learn this distribution in $\mathbb{R}^{512}$, using a five-layer residual ReLU fully connected network (FCN) of width 256.\begin{figure}[H]
\begin{minipage}[t]{0.55\linewidth}
\vspace{0pt}\centering
\includegraphics[width=\linewidth]{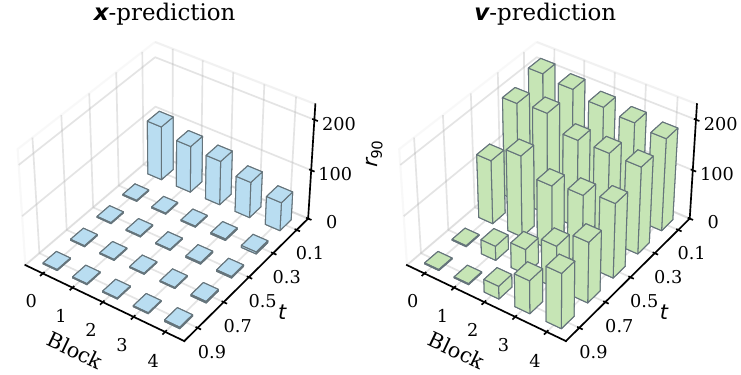}
\caption{\textbf{Toy experiment.} 
Each bar gives the number of directions needed to explain $90\%$ of the hidden-state variance. 
The hidden states under $\vec{x}$-prediction are low-dimensional, whereas those under $\vec{v}$-prediction are not, illustrating the burden. 
}
\label{fig:toy-residual-rank}\label{fig:toy-rank-3d-supp}
\end{minipage}\hfill
\begin{minipage}[t]{0.43\linewidth}
\vspace{0pt}
The generation results are consistent with those reported by \citet{li2026back}: $\vec{x}$-prediction recovers the distribution and $\vec{v}$-prediction fails. Our focus is on the internal states. Figure~\ref{fig:toy-rank-3d-supp} shows that the FCN with $\vec{x}$-prediction concentrates residual-state variance in a few directions except at the highest noise level, while $\vec{v}$-prediction needs over a hundred to explain $90\%$ of the variance by the last block. Even when the input is nearly clean, the FCN with $\vec{v}$-prediction still spreads residual-state variance over many directions, so later layers receive broadly varying states. More details are in Appendix~\ref{app:toy}. \end{minipage}
\end{figure}

This toy experiment makes concrete how the prediction target shapes internal representations in the residual stream. In this paper, we examine this insight with diffusion Transformers on real data. 
\section{Exploitable Low-Dimensionality in Patch Space}\label{sec:analysis}
We first identify the exploitable low-dimensional structure in target patches (Section~\ref{sec:patch-geometry}), then trace the way it shapes input filtering and the resulting intermediate representations (Section~\ref{sec:target-burden}). Finally, we show that a similar low-dimensionality and representation-learning behavior emerge in the main stream of DiT blocks in recent decoupled designs in pixel space with $\vec{v}$-prediction (Section~\ref{sec:architecture-reconciliation}).

\subsection{Spectral Concentration in Patch Space}\label{sec:patch-geometry}\label{sec:whitening}
\begin{figure}[!htbp]
\centering\includegraphics[width=\textwidth]{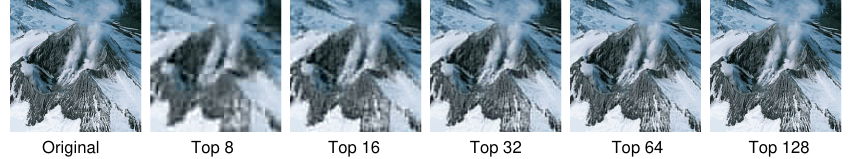}
\par\vspace{-5pt}
\caption{\textbf{A few patch directions preserve recognizable structure.} Each reconstruction keeps the indicated number of principal directions within every $16\times16$ patch. Eight directions preserve the shape of the mountain, and additional directions recover texture.}
\label{fig:patch-reconstruction}
\end{figure}

A DiT processes (noisy) images patch by patch, so the relevant geometry is the \emph{patch geometry}. It is therefore natural to ask whether clean images are low-dimensional in patch space.
For clean $16\times16$ RGB patches in ImageNet $256^2$, just eight of 768 principal directions explain $90\%$ of the variance and preserve recognizable image structure (Figure~\ref{fig:patch-reconstruction}), while the velocity target $\vec v=\vec x-\vec\epsilon$ requires 668 directions to reach the same fraction (Table~\ref{tab:target-geometry} and Appendix~\ref{app:endpoints}).

\begin{table}[H]
\begin{minipage}[c]{0.49\linewidth}
To test the role of spectral concentration, we whiten clean patches by an invertible linear transform that flattens the spectrum while mapping the clean-data manifold to another smoothly equivalent manifold. Training in the whitened space raises $\vec{x}$-prediction FID from $10.19$ to $132.95$, suggesting that spectral concentration in patch space may underlie the success of $\vec{x}$-prediction.
\end{minipage}\hfill
\begin{minipage}[c]{0.5\linewidth}
\centering\small\setlength{\tabcolsep}{3pt}
\begin{tabular}{@{}l>{\columncolor{successcolumn}}c>{\columncolor{failurecolumn}}c>{\columncolor{failurecolumn}}c@{}}
\toprule
Prediction target & Clean & Velocity & Clean \\
Data space & Pixel & Pixel & Whitened \\
\midrule
FID $\downarrow$ & 10.19 & 139.83 & 132.95 \\
Effective rank & 4.80 & 367.37 & 768 \\
Stable rank & 1.41 & 5.94 & 768 \\
$90\%$ variance rank & 8 & 668 & 692 \\
\bottomrule
\end{tabular}
\caption{\textbf{Whitening degrades $\vec{x}$-prediction.}}
\label{tab:target-geometry}
\end{minipage}
\end{table}

\subsection{How Patch Spectral Concentration Shapes Representations}\label{sec:target-burden}\label{sec:representation-evidence}
Following the toy experiment (Figure~\ref{fig:toy-rank-3d-supp}), we now show how patch spectral concentration is reflected in the representations along the residual stream.

\textbf{Spectral concentration induces selective patch filtering.} The optimal velocity estimator is $\E[\vec{v}\mid\vec x_t,t]=(\E[\vec x\mid\vec x_t,t]-\vec x_t)/(1-t)$. Under direct $\vec v$-prediction, the network must reproduce the noisy-input term, so its residual stream must carry corruption-dependent variation even along the many weak directions created by spectral concentration, where $\E[\vec x\mid\vec x_t,t]$ varies little. Under $\vec x$-prediction, the analytic long skip supplies this term, so the backbone only needs $\E[\vec x\mid\vec x_t,t]$, and the patch embedding, where the residual stream begins, can suppress these directions (Appendix~\ref{app:affine-prediction}).

\begin{figure}[!htbp]
\centering\includegraphics[width=\linewidth]{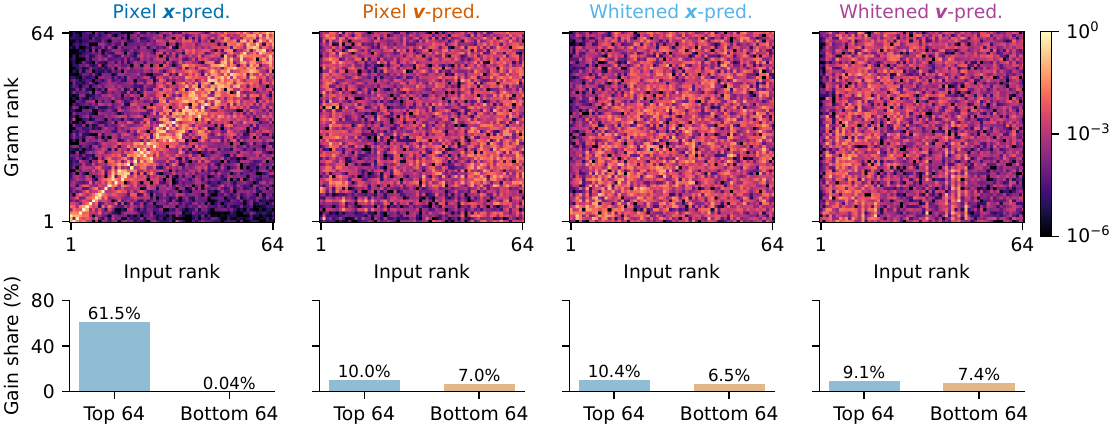}
\par\vspace{-5pt}
\caption{\textbf{Selective filtering depends on target and patch geometry.} Top: squared overlaps between the leading 64 input directions and embedding-Gram eigenvectors; a bright diagonal indicates alignment. Bottom: embedding gain on the leading and trailing 64 directions, normalized over all 768. Whitened axes retain the original clean-PCA order.}
\label{fig:target-embedding}
\end{figure}

To examine this filtering, we measure the directional gain $g_i=\|\mW_{\mathrm{in}}\vec u_i\|^2$, which quantifies how strongly the embedding transmits variation along the clean-patch principal direction $\vec u_i$, and compare the eigenvectors of $\mW_{\mathrm{in}}^\top\mW_{\mathrm{in}}$ with these directions (Appendix~\ref{app:embedding-metrics}). Pixel $\vec x$-prediction assigns $61.5\%$ of total gain to the leading 64 clean-patch directions and only $0.04\%$ to the trailing 64, compared with $10.0\%$ and $7.0\%$ for $\vec v$-prediction (Figure~\ref{fig:target-embedding}).

\begin{figure}[!htbp]
\centering\includegraphics[width=\linewidth]{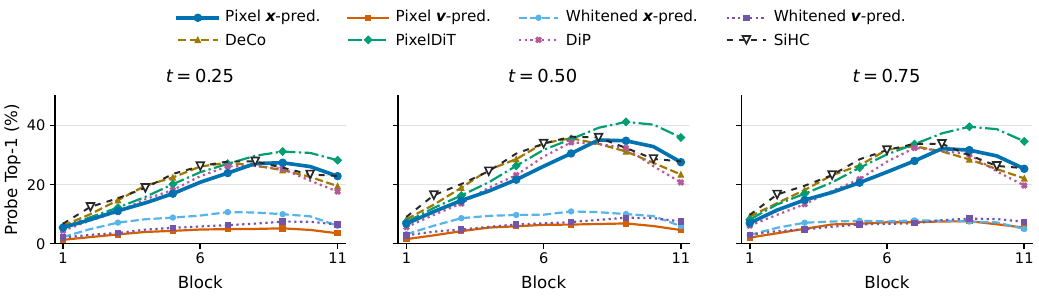}
\par\vspace{-5pt}
\caption{\textbf{Class readability follows the burden.} Linear-probe Top-1 accuracy along the residual stream for the plain DiT in pixel and whitened space, three decoupled $\vec v$-prediction models (Section~\ref{sec:architecture-reconciliation}) and the shared workspace of SiHC (Section~\ref{sec:sihc-method}). Pixel $\vec x$-prediction and the decoupled models are markedly more class-readable than pixel $\vec v$-prediction and both whitened models. Columns fix the clean coefficient $t$, and shading shows the standard deviation across noise draws.}
\label{fig:linear-probes}\label{fig:target-probes}\label{fig:path-probes}
\end{figure}

\textbf{Noise burden and representation quality.}
For two noisy versions $\vec x_t$ and $\vec x'_t$ of the same clean image, $\vec v$-prediction must keep their noise-dependent differences distinguishable in the residual stream, whereas $\vec x$-prediction does not, since the long skip supplies $\vec x_t$. This gives the backbone greater freedom to learn noise-robust representations. Linear probes along the residual stream (Figure~\ref{fig:linear-probes}) show that pixel-space $\vec x$-prediction develops much stronger class features than $\vec v$-prediction, whereas both whitened models remain poorly class-readable, even under $\vec x$-prediction. Together with the embedding measurements, these results connect spectral concentration to selective noise filtering and stronger representations, whose quality is closely associated with generation quality \citep{yu2025repa}.

\subsection{Decoupled Designs Exploit Low-Dimensionality in the Main Stream}\label{sec:architecture-reconciliation}
\begin{figure}[!htbp]
\centering\includegraphics[width=\linewidth]{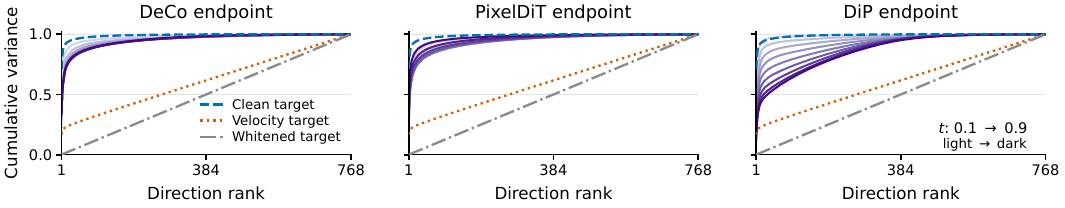}
\par\vspace{-5pt}
\caption{\textbf{Semantic endpoints are far more concentrated than the velocity target.} Target spectra provide a reference for the prediction demand. A faster rise means that fewer directions carry the variance. All three B-size models train successfully under $\vec v$-prediction, reaching FID $6.33$--$13.31$ compared with $139.83$ for the plain DiT (Table~\ref{tab:bsize-decoupled-fid}).}
\label{fig:semantic-endpoints}
\end{figure}

Decoupled architectures offer another way to exploit low-dimensional clean structure while predicting velocity. DeCo, PixelDiT and DiP supply noisy pixels directly to a decoder conditioned on the main Transformer's features (Table~\ref{tab:pixel-paths}), allowing local noise-dependent variation to reach the output through a separate path. We call the features passed to the decoder the \emph{semantic endpoint}, which reflects the residual-stream burden on the main backbone of DiT blocks. As expected, their spectra are also concentrated (see Figure~\ref{fig:semantic-endpoints}). A low-dimensional interface thus emerges naturally through the end-to-end training of the decoupled Transformer designs.

\begin{figure}[!htbp]
\centering\includegraphics[width=\linewidth]{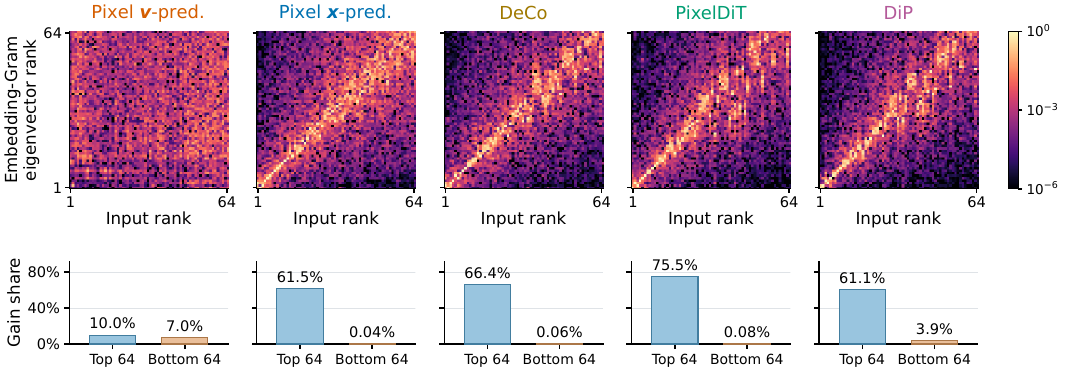}
\par\vspace{-5pt}
\caption{\textbf{Decoupled models filter their input like $\vec x$-prediction.} The top row shows squared overlaps between clean-patch principal directions and embedding-Gram eigenvectors, where a bright diagonal indicates alignment. The bottom row shows the share of input gain on the leading and trailing 64 clean-patch directions.}
\label{fig:embedding-filtering}
\end{figure}

As in a plain DiT with $\vec{x}$-prediction, this structure is reflected at the backbone's entrance and through its depth. Specifically, the patch embeddings align with leading clean-patch directions and suppress trailing directions (Figure~\ref{fig:embedding-filtering}), while intermediate states become much more class-readable than those of plain $\vec v$-prediction (Figure~\ref{fig:linear-probes}). These observations reconcile the decoder-based designs with $\vec x$-prediction: both provide a separate path for noisy input, accompanied by selective filtering and stronger representations in the main stream. Even a zero-initialized scalar skip from the input to the output learns such a path under $\vec v$-prediction, reaching FID $9.99$ at 200 epochs (Appendix~\ref{app:learned-long-skip}).

Together, these results connect patch geometry, prediction targets and architecture through residual-stream burden. We use the output target-patch covariance spectrum as a measurable proxy for \emph{the size} of this burden: it describes how broadly target variation is distributed across directions. By this measure, the velocity target places a far larger burden than the clean target ($r_{90}$ of 668 against 8). The architecture, in turn, determines how this burden is assigned to the main stream and separate paths: with a separate pixel path, the semantic endpoints of decoupled models need only 35 to 309 directions at $t=0.9$ (Appendix~\ref{app:endpoints}). When the burden instead stays on the main stream, as in plain $\vec v$-prediction, the stream itself may need more room. We next test whether expanding the residual stream helps accommodate this broader prediction demand.

\begin{figure}[!t]
\centering\includegraphics[width=\linewidth]{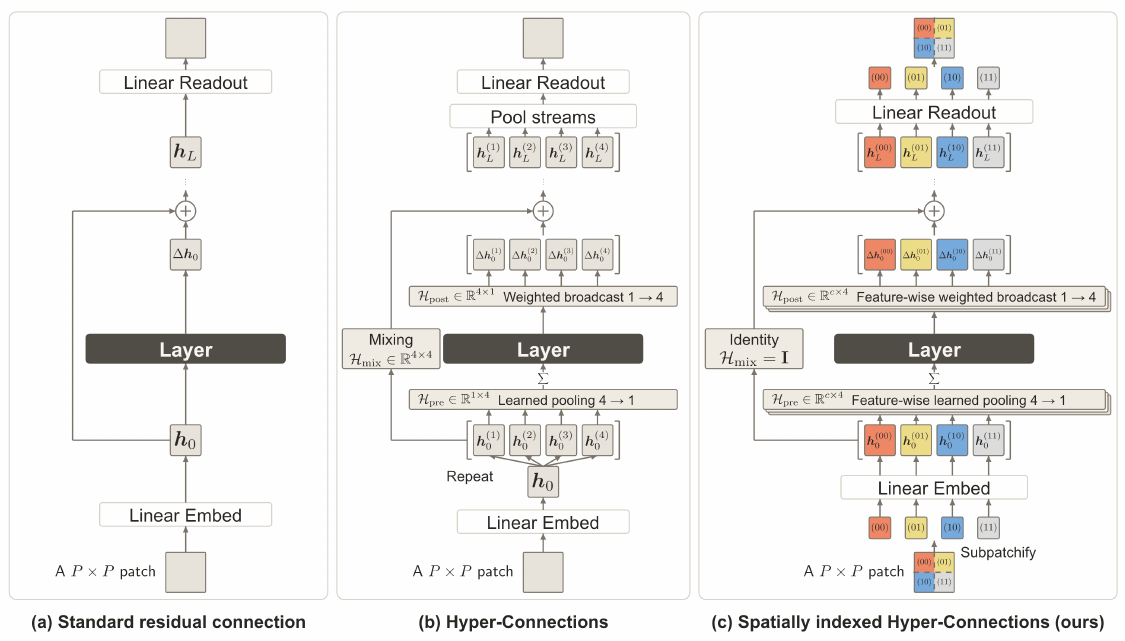}
\par\vspace{-5pt}
\caption{\textbf{Persistent state and workspace in three residual designs,} read from bottom to top. (a) A plain residual state is also the layer workspace. (b) HC carries several states but reads one width-$C$ workspace for each update. (c) SiHC assigns each state a local input and prediction, tracked by color, and connects the states to shared computation through feature-wise read/write maps.}
\label{fig:sihc-method}
\end{figure}

\section{Accommodating Residual-Stream Burden}\label{sec:beyond}\label{sec:spatial-state}\label{sec:bandwidth}
Having shown that separate pixel paths can relieve the main backbone of carrying noise to the final output, we now ask whether expanding the \emph{capacity of the persistent state} lets the backbone accommodate this requirement internally. We call the persistent state's width the \emph{residual-stream bandwidth}. In a plain Transformer, increasing it also widens every attention and MLP computation, since the residual state itself serves as the workspace (Figure~\ref{fig:sihc-method}(a)). Hyper-Connections (HC) separate persistent state from workspace \citep{zhu2025hyperconnections}, allowing us to expand residual-stream bandwidth without widening those computations.

\subsection{Expanding Residual-Stream Bandwidth with Hyper-Connections}\label{sec:hc-probe}
HC replaces $\vec h_\ell$ with $S$ persistent states $\mH_\ell=[\vec h_\ell^{(1)},\ldots,\vec h_\ell^{(S)}]^\top\in\mathbb R^{S\times C}$, from which each update reads one width-$C$ workspace before writing its result back (Figure~\ref{fig:sihc-method}(b)),
\begin{equation}(\vec h_\ell^{\mathrm{in}})^\top=\Hpre^\ell\mH_\ell,\qquad \mH_{\ell+1}=\Hmix^\ell\mH_\ell+\Hpost^\ell(\Delta\vec h_\ell)^\top,\qquad \Delta\vec h_\ell=\mathcal F_\ell(\vec h_\ell^{\mathrm{in}},t).\label{eq:hc}\end{equation}
The read map $\Hpre^\ell\in\mathbb R^{1\times S}$ pools the states, the write map $\Hpost^\ell\in\mathbb R^{S\times1}$ distributes the update, and $\Hmix^\ell\in\mathbb R^{S\times S}$ mixes the carried states. The residual-stream bandwidth becomes $SC$ while the workspace width stays $C$. In dynamic HC, these maps are computed for every token from its current states, so the connections adapt to the input \citep{zhu2025hyperconnections}. In particular, mHC \citep{xie2025mhc}, a dynamic variant that keeps $\Hmix^\ell$ doubly stochastic for stable propagation through depth, has been adopted in open-source LLMs such as DeepSeek-V4 \citep{deepseek2026v4}.

\begin{table}[H]
\begin{minipage}[c]{0.53\linewidth}
The broader target spectrum of $\vec v$-prediction suggests a larger benefit from added bandwidth than for $\vec x$-prediction (Section~\ref{sec:analysis}). We test this prediction in both pixel space and the DINOv2-B latent space used by RAE \citep{zheng2026rae} using four-stream mHC to expand the bandwidth at nearly unchanged parameters and GFLOPs (depth, token count and computational widths remain fixed). With this mHC intervention, $\vec v$-prediction improves substantially: its pixel-space FID falls from $139.8$ to $25.4$ and added bandwidth nearly closes the gap in DINOv2-B space.

\end{minipage}\hfill
\begin{minipage}[c]{0.45\linewidth}
\centering\small\setlength{\tabcolsep}{4pt}
\begin{tabular}{@{}lcc@{}}\toprule
Prediction & Plain FID $\downarrow$ & mHC FID $\downarrow$\\\midrule
Pixel, $\vec x$ & 10.19 & 9.78\\
Pixel, $\vec v$ & 139.83 & 25.36\\\midrule
DINOv2-B, $\vec x$ & 5.50 & 3.93\\
DINOv2-B, $\vec v$ & 16.56 & 3.98\\\bottomrule
\end{tabular}
\caption{\textbf{HC helps $\vec v$-prediction most.} All models are B-size and use Heun-50 with CFG $2.9$. Each pixel patch or DINOv2-B token has 768 coordinates.}
\label{tab:hc-target}\label{tab:dino-b}
\end{minipage}
\end{table}

\subsection{Spatially Indexed Hyper-Connections}\label{sec:sihc-method}
Copied streams expand bandwidth but leave the division of the burden among streams to training. \emph{Spatially Indexed Hyper-Connections} (SiHC) make this division explicit by assigning each persistent state the input and output of one subpatch, while attention/MLP layers read a shared workspace from all states (Figure~\ref{fig:sihc-method}(c)), keeping the number of workspace tokens unchanged.
\begin{enumerate}[leftmargin=*,nosep,label=(\arabic*)]
\item \textbf{Local states carry local predictions.} Each $P\times P$ computational patch is divided into $S=(P/b)^2$ subpatches. State $s$ embeds only its noisy subpatch, $\vec h_0^{(s)}=\mW_{\mathrm{in}}\vec p_t^{(s)}$, and predicts its local velocity, $\widehat{\vec v}^{(s)}=\mW_{\mathrm{out}}\vec h_{\mathrm{final}}^{(s)}$. This reduces each state's prediction responsibility from $3P^2$ to $3b^2$ pixel coordinates, while the $S$ local predictions tile the full velocity patch.
\item \textbf{Identity carry keeps local detail available.} With $\Hmix^\ell=\mI_S$, each state retains its own carry path independently of its contribution to the current workspace.
\item \textbf{Static feature-wise access connects local states to shared computation.} Learned, input-independent maps $\Hpre^\ell,\Hpost^\ell\in\mathbb R^{C\times S}$, shared across spatial cells, let each channel read its own combination of states and distribute its update with its own weights.
\end{enumerate}
For channel $c$, the read and write around each attention or MLP update are
\begin{equation}
h_{\ell,c}^{\mathrm{in}}=\textstyle\sum_s\Hpre^\ell[c,s]\,h_{\ell,c}^{(s)},\qquad
h_{\ell+1,c}^{(s)}=h_{\ell,c}^{(s)}+\Hpost^\ell[c,s]\,\Delta h_{\ell,c}.
\label{eq:sihc-read}
\end{equation}
Read-back makes local information available to subsequent computation, and write-back distributes the resulting updates among local prediction states. Dynamic maps must be recomputed from the full $S$-stream state before every update, which requires materializing the wide state at every sublayer. With static maps and identity carry, reads and writes instead compose across a stage, allowing exact kernel fusion that materializes the wide state only at stage boundaries. This late materialization reduces the memory traffic and footprint of the expanded state, lowering peak training memory from $30.18$ to $19.82$ GiB for the sixteen-stream B model (Appendix~\ref{app:fusion}).

\textbf{Ablation study.} Table~\ref{tab:progression} tests these choices in sequence on a B-size backbone with $16\times16$ computational patches and direct $\vec v$-prediction. Assigning each of four streams its own $8\times8$ input and prediction region lowers FID from $25.4$ to $13.2$ at the same bandwidth, demonstrating the value of smaller local responsibilities. Static scalar maps with identity carry improve FID further, and channel-specific read/write maps refine access to the local states without input-dependent connections. Dividing the patch into sixteen $4\times4$ regions then expands bandwidth while reducing each state's prediction responsibility to 48 pixel coordinates, bringing FID to $8.07$, within the range of decoupled B-size models (Table~\ref{tab:bsize-decoupled-fid}).

\begin{table}[!htbp]
\centering\small\setlength{\tabcolsep}{4pt}
\caption{\textbf{Each design choice improves $\vec v$-prediction.} Parameters and GFLOPs remain nearly unchanged across the progression.}
\label{tab:progression}
\begin{tabular*}{\linewidth}{@{\extracolsep{\fill}}llccc@{}}
\toprule
Design & What changes & State region & Streams & FID $\downarrow$\\
\midrule
Plain DiT & & $16\times16$ & 1 & 139.83\\
mHC, copied input & Four persistent states & $16\times16$ & 4 & 25.36\\
mHC, spatial input & Local input and prediction per state & $8\times8$ & 4 & 13.25\\
\quad + identity carry, scalar maps & Static maps without state mixing & $8\times8$ & 4 & 11.50\\
\quad + feature-wise maps (SiHC) & Channel-specific read and write & $8\times8$ & 4 & 10.10\\
\quad + $4\times4$ subpatches & More states, smaller regions & $4\times4$ & 16 & 8.07\\
\bottomrule
\end{tabular*}
\end{table}

The sublayerwise SiHC workspace also develops much stronger class features than plain $\vec v$-prediction, tracking pixel-space $\vec x$-prediction through depth (Figure~\ref{fig:linear-probes}). Organizing persistent state and its access thus recovers class-readable features without a separate pixel decoder. Further noise-sensitivity measurements appear in Appendix~\ref{app:sihc-sensitivity}.

This design also scales beyond the B-size analysis (Appendix~\ref{app:sihc-scaling}). With REPA, SiHC-XL reaches FID $1.71$ (Table~\ref{tab:main-performance}), competitive with strong pixel-space models while retaining linear read/write maps in place of specialized decoders or cross-scale attention.

\begin{table}[!htbp]
\centering\small
\caption{\textbf{Pixel-space generation on ImageNet-$256^2$.} We compare with a selection of recent pixel-space models under a common sampling setting: all rows use Heun-50, and our evaluations follow JiT's FID protocol \citep{li2026back}. Some of these models achieve better results with their own sampling configurations, for example DeCo and PixelDiT. These results and further details are given in Appendix~\ref{app:performance-context}. $^{\dagger}$Uses REPA; --: not reported.}
\label{tab:main-performance}
\begin{tabular*}{\linewidth}{@{\extracolsep{\fill}}lccccc@{}}
\toprule
Model & Prediction & Params (M) & GFLOPs & FID $\downarrow$ & IS $\uparrow$\\
\midrule
JiT-H/16 \citep{li2026back} & $\vec x$ & 953 & 364 & 1.86 & 303.4\\
JiT-G/16 \citep{li2026back} & $\vec x$ & 2000 & 767 & 1.82 & 292.6\\
PixelREPA-H/16 \citep{shin2026pixelrepa} & $\vec x$ & 953 & 364 & 1.81 & 317.2\\
PixNerd-XL/16$^{\dagger}$ \citep{wang2026pixnerd} & $\vec v$ & 700 & 268 & 1.89 & 309.0\\
DiP-XL/16$^{\dagger}$ \citep{chen2026dip} & $\vec v$ & 631 & 255 & 1.75 & 273.0\\
DeCo-XL/16$^{\dagger}$ \citep{ma2026deco} & $\vec v$ & 682 & 245 & 1.69 & 304.0\\
PixelDiT-XL$^{\dagger}$ \citep{yu2026pixeldit} & $\vec v$ & 797 & 311 & 1.69 & 292.0\\
AsymFlow-H/16$^{\dagger}$ \citep{chen2026asymflow} & $\vec x-\mP\vec\epsilon$ & 953 & 363 & 1.57 & --\\
\midrule
SiHC-XL$^{\dagger}$ (Ours) & $\vec v$ & 763 & 299 & 1.71 & 299.6\\
\bottomrule
\end{tabular*}
\end{table}

\section{Discussion, Limitations and Future Work}\label{sec:discussion}

The high-level view of this work is to treat a neural network not as a function from noisy input to target but as a system whose persistent state must serve both prediction and representation learning. From this view, residual-stream burden explains a central difficulty in training diffusion models: predicting a noisy target requires the residual stream to keep the noise, and this induces friction with representation learning. Analyzing the patch space provides proxies for the size of the burden and shows why the issue becomes especially severe for high-dimensional data and large patches, where the noise spreads over far more directions. Through this lens, we reconcile recent progress in pixel-space diffusion, where changing the prediction parameterization and changing the architecture succeed for a common reason, namely relieving the main stream of the noise. Building on this understanding, we explore a complementary direction that accommodates the burden by expanding residual-stream bandwidth, with SiHC serving as an algorithmic probe.

\textbf{Limitations.} Our analysis shows that residual-stream burden shapes the learned representations, but it does not trace how this shaping unfolds during training, which is fundamentally difficult. Representations in deep networks emerge from gradient dynamics, in which representation learning and gradient geometry are coupled. The gradient of a hidden unit combines its inputs, weighted by backpropagated errors, and these inputs are themselves learned by earlier layers, so each representation update also changes the geometry of subsequent updates. This coupling prevents us from tracing how specific representations form, and it also leaves open how much bandwidth representation learning itself requires. Architecture and data geometry both act by reshaping the network parameters through these dynamics, and existing theoretical tools remain insufficient to characterize the resulting \emph{implicit bias} precisely (Appendix~\ref{app:implicit-bias}).

\textbf{Future directions.} Viewing residual-stream burden through its friction with representation learning raises several questions.
\begin{enumerate}[leftmargin=*,nosep,label=(\arabic*)]
\item \emph{How does the implicit bias of gradient descent interact with friction?} Selective filtering, concentrated semantic endpoints and a learned long skip (Appendix~\ref{app:learned-long-skip}) all emerge without an explicit objective. We conjecture that when the architecture offers a path that carries the burden outside feature-building computation, gradient descent drifts toward solutions with less friction, consistent with a preference for simpler solutions. Testing this conjecture requires a way to characterize friction during training, for example as interference between the gradients that preserve the target and those that build features.
\item \emph{Can residual-stream burden benefit representation learning?} When the target is itself a pretrained representation, as in RAE \citep{zheng2026rae, singh2026raev2}, the burden consists of features that later layers can build on and may even facilitate representation learning, offering a possible reason for the fast convergence of RAE.
This also invites studying \emph{diffusability}, the question of which latent space suits diffusion models \citep{skorokhodov2025diffusability,yao2025lightningdit,leng2025repae}, from the perspective of neural network training.
\item \emph{When do generation and understanding reinforce each other?} The same training perspective extends to unified models of generation and visual understanding \citep{wu2025janus,hu2026orion,wang2026uniddt}. Recent unified models that generate within a shared semantic latent space report improved alignment between the two tasks \citep{jin2026latentum}. Characterizing the burden that generation places on shared representations, and when it is compatible with understanding, may help answer this question.
\end{enumerate}
\par
\paragraph{AI Use Statement.}
Generative AI tools were used to assist with language editing and improve the readability of the manuscript, and to prepare preliminary versions of figures and visualizations. Codex and Claude Code also assisted with code development and debugging, including the implementation of kernel fusion and the selection of tiling configurations. All AI-assisted text, figures, and code were reviewed, revised as needed, and verified by the authors. The authors formulated the main research ideas, made the methodological and experimental decisions, and are responsible for the interpretation of results and the scientific conclusions. The authors take full responsibility for the final content of this work.

\paragraph{Reproducibility Statement.}
The supplementary materials include anonymous SiHC source code with model configurations, training and evaluation entry points, fused kernels, benchmark utilities, correctness tests and a toy-analysis suite. The appendices document the data transformations and spectral measurements (Appendix~\ref{app:pca-data}), representation diagnostics (Appendices~\ref{app:sensitivity} and~\ref{app:probes}), toy setup (Appendix~\ref{app:toy}), training and sampling protocols (Appendix~\ref{app:sihc-config}), and the read/write implementation and fusion benchmarks (Appendices~\ref{app:grad} and~\ref{app:fusion}).

\bibliographystyle{plainnat}
\bibliography{references}
\clearpage

\appendix
\section{Additional Related Work}\label{sec:rw}\label{app:related}

\subsection{Diffusion Models and Prediction Targets}\label{app:prediction-targets}
Noise, clean-data and velocity prediction describe the same diffusion process through interconvertible outputs and time-dependent loss weights \citep{ho2020ddpm,salimans2022progressive,lipman2023flow,liu2023rectified}, but they can assign different prediction responsibilities to the network. For example, clean prediction under a velocity-space objective supplies the noisy-input term through an analytic skip, changing what the residual stream must preserve \citep{li2026back}. Recent studies examine the resulting learning differences through prediction dimensionality, covariance anisotropy and pixel-space training \citep{jin2026kdiff,fu2026jlt,jiang2026pixeltraining}. We connect these differences to learned input filtering and the representations available to subsequent layers (Sections~\ref{sec:target-burden} and~\ref{sec:architecture-reconciliation}).

\subsection{Latent Diffusion and Pixel-Space Modeling}\label{app:latent-pixel}
Latent diffusion reduces denoising cost through autoencoder compression \citep{rombach2022ldm}, enabling scalable Transformer generators such as DiT and SiT \citep{peebles2023dit,ma2024sit}. This efficiency also makes generation depend on the representation supplied by the autoencoder: reconstruction accuracy alone does not determine how easily diffusion can learn the latent distribution \citep{yao2025lightningdit,skorokhodov2025diffusability}. Aligning latents with visual features, jointly tuning the autoencoder and denoiser, and generating directly in pretrained representation spaces address this interaction from complementary directions \citep{yao2025lightningdit,leng2025repae,zheng2026rae}.

\paragraph{Why pixel space?}
Pixel-space modeling pursues a self-contained image synthesis pipeline in which the generative objective shapes the full path to image detail, without a separately pretrained image tokenizer. End-to-end diffusion and autoregressive modeling pursue this integration \citep{hoogeboom2023simple,tschannen2025jetformer}. Learning the image synthesis path jointly also avoids fixing the compression--reconstruction trade-off before training the generator \citep{rombach2022ldm,chen2025dcae}. Recent pixel-space architectures make this integration practical by separating coarse computation from detailed prediction \citep{chen2026dip,yu2026pixeldit}. Large patches reduce attention cost but assign many pixel coordinates to each residual stream, making this setting useful for studying how prediction requirements interact with representation learning.

\subsection{Architecture Design in Diffusion Models}\label{app:diffusion-architectures}
\label{app:pixel-decoders}
Diffusion architectures distribute global modeling and detailed prediction through hierarchy or decoupled interfaces. U-Net, hourglass and staged designs allocate computation across resolutions \citep{dhariwal2021adm,hoogeboom2025sid2,nawrot2022hourglass,crowson2024hdit,chen2025pixelflow,guo2026pixelu}, whereas decoupled designs assign detailed prediction to a head conditioned on backbone features, in both latent and pixel space \citep{wang2026ddt,zheng2026rae,chen2026dip,ma2026deco,yu2026pixeldit,wang2026pixnerd}. Analytic input skips induced by prediction reparameterization provide another way to supply local information outside the main stream \citep{li2026back,chen2026asymflow}. The common design principle is to preserve access to detail without requiring all of it to pass through the shared computation.

\label{app:hyperdit}
The interface determines how that detail remains accessible: decoder-based designs use final backbone features, while cross-attention connects data or fine-grained states to latent or coarse computation \citep{jabri2023rin,he2026hyperdit,bai2026duspit}. SiHC uses persistent local states with feature-wise linear access, keeping both state and workspace resolutions fixed through depth (Table~\ref{tab:taxonomy}). Our analysis connects these choices to prediction requirements: even at fixed granularity, the target and patch geometry change the burden on the residual stream (Section~\ref{sec:architecture-reconciliation}).

\begin{table}[!htbp]
\centering\small
\caption{\textbf{Prediction interfaces in pixel-space models.} Analytic skips, conditioned decoders and cross-attention provide different paths from noisy inputs to predictions. SiHC connects persistent local states to shared computation through linear read/write maps.}
\label{tab:taxonomy}
\setlength{\tabcolsep}{4pt}
\renewcommand{\arraystretch}{1.08}
\begin{tabularx}{\linewidth}{@{}>{\raggedright\arraybackslash}p{0.17\linewidth}*{3}{>{\raggedright\arraybackslash}X}@{}}
\toprule
Model & Noisy-input path & Connection to backbone & Prediction computation \\
\midrule
\multicolumn{4}{@{}l}{\textit{Analytic output paths}}\\
JiT \newline{\scriptsize\citep{li2026back}} & Noisy input enters both the patch embedding and an analytic output skip. & Final patch features predict the clean image. & Linear patch head, then analytic conversion to velocity. \\
\addlinespace
AsymFlow \newline{\scriptsize\citep{chen2026asymflow}} & An analytic input path recovers velocity outside a chosen noise subspace. & A JiT backbone predicts full-dimensional data minus low-rank noise. & Patch head with projection-based analytic velocity recovery. \\
\midrule
\multicolumn{4}{@{}l}{\textit{Decoders conditioned on final backbone features}}\\
DiP \newline{\scriptsize\citep{chen2026dip}} & Noisy RGB patches enter local U-Nets. & Context tokens enter U-Net bottlenecks. & Patch-local convolutions with down/up blocks and skips. \\
\addlinespace
DeCo \newline{\scriptsize\citep{ma2026deco}} & Noisy RGB and positions form pixel queries. & Upsampled coarse features supply AdaLN-zero conditioning. & Pixel-wise residual MLPs without attention. \\
\addlinespace
PixelDiT \newline{\scriptsize\citep{yu2026pixeldit}} & Noisy pixels enter narrow tokens. & Patch tokens supply per-pixel AdaLN parameters. & Global attention on compressed tokens, then expansion and pixel MLPs. \\
\addlinespace
PixNerd \newline{\scriptsize\citep{wang2026pixnerd}} & Noisy RGB and coordinates query local neural fields. & Patch tokens predict normalized MLP weights. & Neural-field MLP and output projection per pixel. \\
\midrule
\multicolumn{4}{@{}l}{\textit{Cross-attention interfaces across depth}}\\
RIN \newline{\scriptsize\citep{jabri2023rin}} & Noisy data are embedded as interface tokens. & Cross-attention reads into latent tokens and writes back to interface tokens. & Self-attention in latent tokens, with output decoded from the interface. \\
\addlinespace
HyperDiT \newline{\scriptsize\citep{he2026hyperdit}} & Separate embeddings initialize coarse semantic and fine-grained streams. & Fine tokens query semantic anchors from multiple backbone depths via cross-attention. & Hyper-Connector blocks refine the fine stream with scale-aware positional encoding. \\
\addlinespace
DuSPiT \newline{\scriptsize\citep{bai2026duspit}} & Parallel patch and subpatch embeddings initialize two branches. & Each subpatch token attends only to its corresponding patch token, using intermediate patch-branch features. & Cross-attention and feed-forward updates in the pixel branch, followed by a clean-image head. \\
\midrule
\multicolumn{4}{@{}l}{\textit{Linear access to persistent local states}}\\
SiHC (ours) & Subpatch states persist with identity carry. & Feature-wise linear reads/writes connect states to shared sublayers. & Local output projections, with attention/MLPs in the shared workspace. \\
\bottomrule
\end{tabularx}
\end{table}

\FloatBarrier
\subsection{Implicit Bias of Gradient Descent}\label{app:implicit-bias}
The filtering patterns in Sections~\ref{sec:target-burden} and~\ref{sec:architecture-reconciliation} can be viewed as an expression of implicit bias: training selects patch directions without an explicit regularizer prescribing that selection, and the directions it retains change with the target and output path. This perspective follows the question raised by \citet{zhang2017rethinking}: when the same network can fit random labels yet generalize on natural data, which properties of training favor useful solutions? For diffusion, the question extends inside the network, because matching the output target leaves open how the model distributes information and computation across its layers.

Existing theory makes parts of this selection process precise under tractable assumptions. Analyses of linear models characterize how the loss, parameterization and optimization geometry induce margin or norm preferences \citep{soudry2018implicit,gunasekar2018geometry,gunasekar2018convolutional}. For nonlinear networks, mean-field descriptions track how features evolve during training, while two-layer analyses establish advantages from learning task-relevant directions \citep{mei2019meanfield,damian2022representations}.

A complementary approach studies which solutions can remain stable under finite-step training, using the step size to constrain the solutions without tracking the full optimization trajectory \citep{wu2018stability,wu2023stability}. The empirical edge-of-stability phenomenon and its self-stabilization mechanism connect this perspective to observed training dynamics \citep{cohen2021edge,damian2023selfstabilization}. For shallow ReLU networks, stability constraints translate into function-space regularity \citep{mulayoff2021stability,nacson2023stability} and generalization guarantees in univariate regression \citep{qiao2024stable}. Their extension to multivariate inputs reveals why dimension and data geometry matter \citep{liang2025shattering,liang2026datageometry}, while sparse connectivity changes the relevant geometry from whole inputs to local patches \citep{liang2026sparseconnectivity}. This line of work makes explicit how architecture and data jointly determine the regularization supplied by training.

The central difficulty is that the geometry driving optimization is itself learned. Even in a two-layer ReLU network, the gradient of a hidden neuron's input weights combines input vectors weighted by prediction errors, output weights and activation gates \citep{liang2026sparseconnectivity}. As those weights and gates change, so do the directions favored by subsequent updates. In a deep network, the inputs to later layers also move as earlier representations change, while backpropagated gradients depend on the computation built above them. Representation learning therefore changes the optimization problem experienced by each layer throughout training. Knowing which information an optimal predictor needs is insufficient to determine which internal representations gradient descent will construct to carry it.

Our experiments give this coupled problem a concrete empirical form. Changing the target alters which patch directions the embedding retains, and changing the output path can recover selective filtering under velocity prediction. Expanding residual-stream bandwidth changes how much information can remain available alongside the shared computation. A theory of these observations would need to explain how target-induced gradients select input directions and how their retention affects feature learning through depth. Residual-stream burden organizes these questions around the information that must remain accessible, linking the observed filtering and bandwidth effects to the optimization challenge discussed in Section~\ref{sec:discussion}.

\subsection{Generation Performance and Training Context}\label{app:performance-context}
Table~\ref{tab:full-performance} extends the main pixel-space comparison to latent models. REPA use is marked in the model names.\footnote{DiP's released XL configuration enables its REPA trainer: \url{https://github.com/NJU-PCALab/DiP/blob/main/configs_c2i/dip_xl.yaml}.} PixNerd's listed FID 1.93 uses 320 training epochs and Euler-100, while DiP's 1.79 uses 600 epochs and Euler-100 \citep{wang2026pixnerd,chen2026dip}.

\paragraph{Sampling configurations.}
Table~\ref{tab:main-performance} uses Heun-50 throughout. We evaluate the released PixNerd, DiP and PixelDiT models, together with SiHC, using JiT's FID protocol on 50,000 samples. Their FID/IS pairs are $1.89/309.0$, $1.75/273.0$, $1.69/292.0$ and $1.71/299.6$, respectively. JiT, PixelREPA and DeCo use their reported Heun-50 results. AsymFlow's reported FID $1.57$ uses ADM evaluation, and its corresponding IS is not reported \citep{chen2026asymflow}. Its asymmetric target is written as $\vec x-\mP\vec\epsilon$ in our time convention, where $\mP$ is a rank-eight patch projector.

Table~\ref{tab:full-performance} retains the broader comparison under each source's published sampling configuration. DeCo's FID $1.62$ uses Euler-250 at 800 epochs \citep{ma2026deco}. PixelDiT's FID $1.61$ uses FlowDPMSolver-100 at 320 epochs, with guidance scale $2.75$ over $[0.10,0.90]$ \citep{yu2026pixeldit}. RAEv2 uses its DINOv3-L $K=7$ configuration with REPA guidance \citep{singh2026raev2}. GFLOPs measure one denoiser forward pass, while total sampling compute also depends on the solver, step count and classifier-free guidance.

\begin{table}[!htbp]
\centering\small\setlength{\tabcolsep}{7pt}
\caption{\textbf{Generation quality and model cost on ImageNet-$256^2$.} Reported results span pixel space and latent spaces based on VAEs or RAEs. GFLOPs count one denoiser forward pass; training and sampling schedules follow each source. REPA is indicated explicitly for the pixel models that use it. --: not reported. $^*$Computed from the published or released configuration (Appendix~\ref{app:compute}).}
\label{tab:full-performance}
\begin{tabular}{@{}lrrrr@{}}
\toprule
Model & Params (M) & GFLOPs & FID $\downarrow$ & IS $\uparrow$\\
\midrule
\multicolumn{5}{@{}l}{\textit{Latent diffusion with VAE}}\\
DiT-XL/2 \citep{peebles2023dit} & 675 & 238 & 2.27 & 278.2\\
SiT-XL/2 \citep{ma2024sit} & 675 & 238 & 2.06 & 277.5\\
SiT-XL/2 + REPA \citep{yu2025repa} & 675 & 238 & 1.42 & 305.7\\
LightningDiT-XL/2 \citep{yao2025lightningdit} & 675 & 238 & 1.35 & 295.3\\
DDT-XL/2 \citep{wang2026ddt} & 675 & 238 & 1.26 & 310.6\\
\midrule
\multicolumn{5}{@{}l}{\textit{Latent diffusion with RAE}}\\
RAE, DiT-XL \citep{zheng2026rae} & 676 & 238 & 1.41 & 309.4\\
RAE, DiT$^{\mathrm{DH}}$-XL \citep{zheng2026rae} & 839 & 323 & 1.13 & 262.6\\
RAEv2, DiT$^{\mathrm{DH}}$-XL \citep{singh2026raev2} & 875$^*$ & 475$^*$ & 1.06 & 255.3\\
\midrule
\multicolumn{5}{@{}l}{\textit{Pixel-space diffusion}}\\
PixelFlow-XL/4 \citep{chen2025pixelflow} & 677 & 5818 & 1.98 & 282.1\\
PixNerd-XL/16 + REPA \citep{wang2026pixnerd} & 700 & 268 & 1.93 & 298.0\\
JiT-H/16 \citep{li2026back} & 953 & 364 & 1.86 & 303.4\\
JiT-G/16 \citep{li2026back} & 2000 & 767 & 1.82 & 292.6\\
PixelREPA-H/16 \citep{shin2026pixelrepa} & 953 & 364 & 1.81 & 317.2\\
DiP-XL/16 + REPA \citep{chen2026dip} & 631 & 255 & 1.79 & 281.9\\
DeCo-XL/16 + REPA \citep{ma2026deco} & 682 & 245 & 1.62 & 301.0\\
PixelDiT-XL + REPA \citep{yu2026pixeldit} & 797 & 311 & 1.61 & 292.7\\
HyperDiT-XL + REPA \citep{he2026hyperdit} & 676 & 551$^*$ & 1.63 & 304.2\\
AsymFlow-H/16 + REPA \citep{chen2026asymflow} & 953 & 363 & 1.57 & --\\
\midrule
SiHC-XL + REPA & 763 & 299 & 1.71 & 299.6\\
\bottomrule
\end{tabular}
\end{table}

\FloatBarrier
\section{Patch Spectrum and Embedding Details}
\label{app:pca-data}
\subsection{Definitions}
For the input embedding, stack the noisy image patches as rows:
\begin{equation}
\mathbf{H}^{0}
=
\mathbf{P}_t \mW_{\mathrm{in}}^\top+\mathbf{B},
\qquad
\mathbf{P}_t=\operatorname{Patchify}(\vec{x}_t),
\label{eq:patch-input-interface}
\end{equation}

Here $\mW_{\mathrm{in}}\in\mathbb R^{C\times d}$ maps each patch vector to a width-$C$ state, and $\mathbf B$ collects additive bias and positional terms. The centered clean-patch covariance is $\bm{\Sigma}_p=\Cov(\vec p)$, with eigendecomposition
\begin{equation}
\bm{\Sigma}_p=
\mU\bm{\Lambda} \mU^\top,
\qquad
\bm{\Lambda}=\operatorname{diag}(\lambda_1,\ldots,\lambda_d),
\qquad
\lambda_1\geq\cdots\geq\lambda_d.
\label{eq:patch-pca}
\end{equation}

For covariance eigenvalues $\lambda_1\ge\cdots\ge\lambda_d\ge0$, let $p_i=\lambda_i/\sum_j\lambda_j$. We use
\begin{equation}
 \er=\exp\!\left(-\sum_i p_i\log p_i\right),\qquad
 \mathrm{SR}=\frac{\sum_i\lambda_i}{\lambda_1},\qquad
 r_\alpha=\min\left\{k:\frac{\sum_{i=1}^k\lambda_i}{\sum_i\lambda_i}\ge\alpha\right\}.
\end{equation}
These statistics characterize the variance spectrum. Under independent isotropic corruption, the noisy patch covariance is
\begin{equation}
\operatorname{Cov}(\vec{p}_t)=t^2\bm{\Sigma}_p+(1-t)^2\mI.
\label{eq:noisy-patch-cov}
\end{equation}
For the covariance of clean $16\times16$ patches, clean ER is $4.796$ and the leading principal component explains $71.13\%$ of variance. At $t=0.5$, noisy-patch ER is $367.371$ and $r_{90}=668$. The raw velocity-target covariance and its rank are derived in Appendix~\ref{app:endpoints}.

\subsection{Patch spectra and additional visual measurements}
\begin{figure}[!htbp]
\centering
\includegraphics[width=\linewidth]{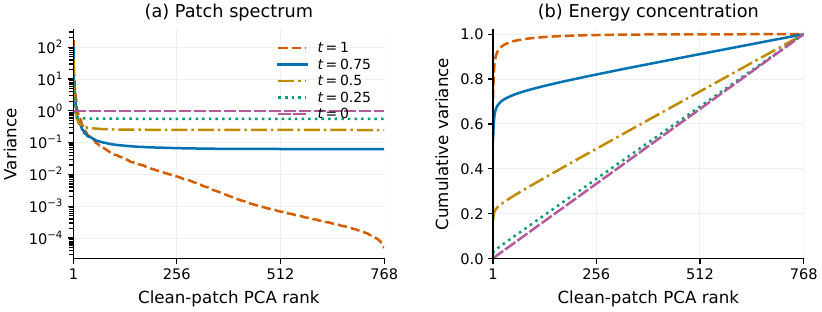}
\caption{\textbf{Corruption spreads patch variance across directions.} Left: variance along clean-PCA directions. Right: cumulative variance. Decreasing the clean coefficient $t$ flattens the spectrum and slows the cumulative rise, so more directions are needed to explain the same fraction of patch variance.}
\label{fig:patch-geometry}
\end{figure}

\begin{figure}[!htbp]
\centering\includegraphics[width=\linewidth]{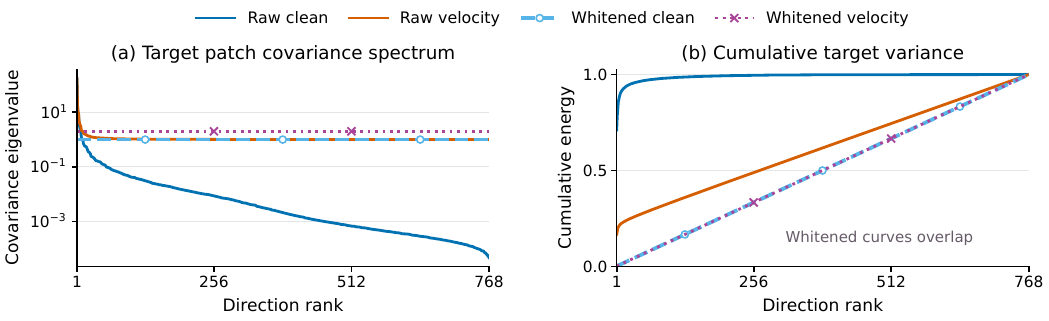}
\par\vspace{-5pt}
\caption{\textbf{Whitening broadens both prediction targets.} Left: target covariance eigenvalues. Right: cumulative variance. The whitened spectra assign equal variance to every direction, so their normalized cumulative curves coincide. Raw clean patches concentrate variance in a small leading subspace.}
\label{fig:target-spectra-supp}
\end{figure}

The covariance and reconstruction basis use 100k ImageNet training images and $16\times16$ RGB patches. Figure~\ref{fig:patch-geometry} shows how corruption flattens the patch spectrum, and Figure~\ref{fig:target-spectra-supp} compares the clean and velocity target spectra in pixel and whitened space.
\begin{remark}[JiT's bottleneck embedding]
\citet{li2026back} use a low-rank linear patch embedding in their ``Just image Transformers'' (JiT). JiT-B/64 successfully generates from $64\times64$ RGB patches with 12,288 coordinates using a bottleneck width of 128. Our learned-filtering results support this design: the fixed rank encourages the embedding to suppress noise-dominated directions before they enter the residual stream.
\end{remark}

\begin{figure}[!htbp]\centering
\includegraphics[width=0.65\linewidth]{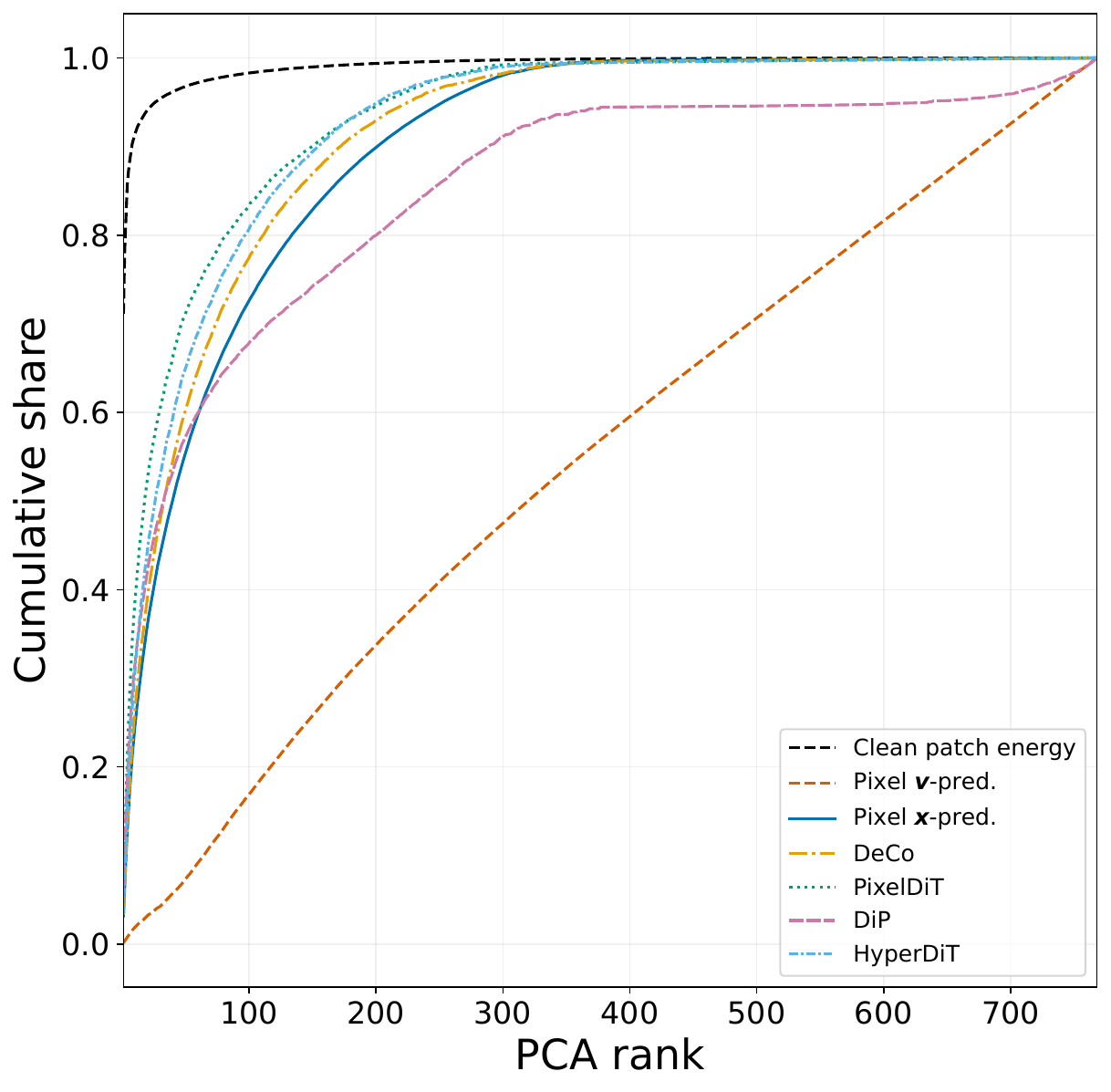}
\caption{\textbf{Cumulative gain shows which patch directions enter the stream.} Each curve sums embedding gain in clean-PCA order. A faster rise means that more gain is assigned to the leading clean directions.}
\label{fig:embedding-cumulative-supp}
\end{figure}
\begin{figure}[!htbp]\centering
\includegraphics[width=0.75\linewidth]{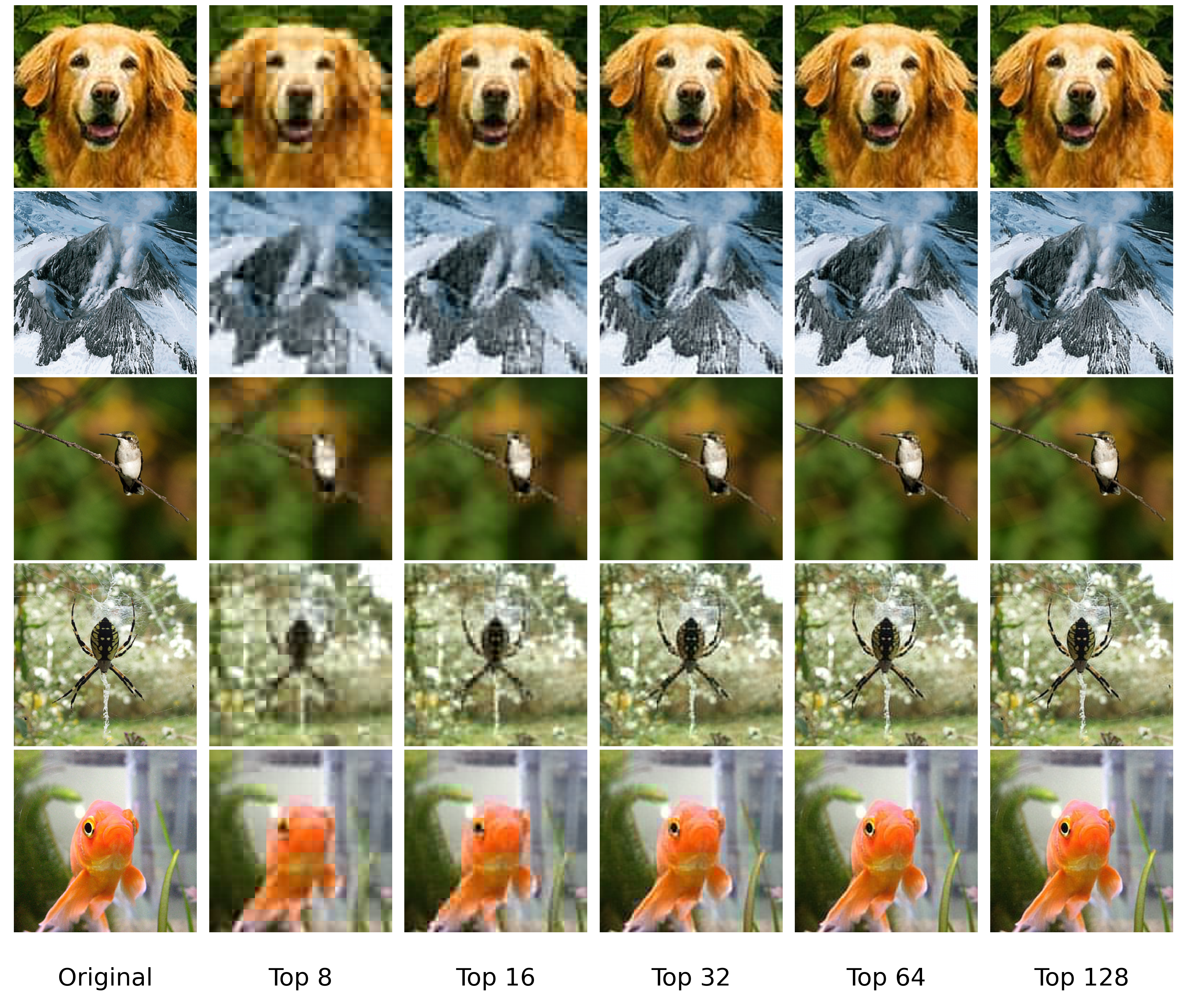}
\caption{\textbf{Leading patch directions preserve recognizable structure across images.} Each row is one image. Columns show the original and reconstructions retaining 8, 16, 32, 64 and 128 principal directions within each $16\times16$ patch. More directions recover fine texture.}
\label{fig:pca-reconstructions}
\end{figure}

Figure~\ref{fig:embedding-cumulative-supp} complements the alignment and head/tail views in the main text, and Figure~\ref{fig:pca-reconstructions} extends the reconstruction in Figure~\ref{fig:patch-reconstruction} to five images.
\subsection{Spectral concentration in video tubelets}
\label{app:video-patch-spectrum}
\begin{figure}[!htbp]
\centering
\includegraphics[width=\linewidth]{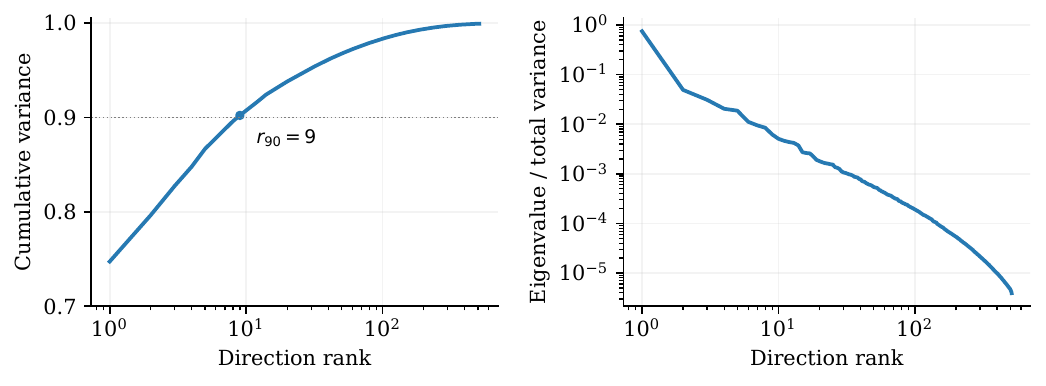}
\par\smallskip
{\small\setlength{\tabcolsep}{7pt}
\begin{tabular}{@{}lrrrr@{}}
\toprule
Tubelet & Dimension & $r_{90}$ & $r_{95}$ & $r_{99}$\\
\midrule
UCF101, $4\times16\times16$ & 3,072 & 9 & 29 & 151\\
\bottomrule
\end{tabular}}
\caption{\textbf{Video tubelets have concentrated spectra.} Left: cumulative variance, with the marker at $r_{90}$. Right: covariance eigenvalues normalized by total variance.}
\label{fig:video-patch-spectrum}
\end{figure}

Spectral concentration is also present in natural video. We sample one RGB tubelet of size $4\times16\times16$ (time $\times$ height $\times$ width) from each of 5,050 UCF101 videos, with 50 videos per class \citep{soomro2012ucf101}, taken from $16\times256\times256$ clips. Each tubelet has 3,072 coordinates with RGB values in $[-1,1]$, and the covariance is centered by the global sample mean. Only nine of the 3,072 principal directions explain $90\%$ of the variance (Figure~\ref{fig:video-patch-spectrum}), compared with eight of 768 for the $16\times16$ image patches in Section~\ref{sec:patch-geometry}. Local spatiotemporal patches of natural video thus concentrate their variance in a few directions, as image patches do.

\subsection{Best-affine prediction and directional gain}\label{app:affine-prediction}
The conditional expectations in Section~\ref{sec:target-burden} describe optimal predictions under squared loss. Restricting to affine maps gives an explicit second-moment calculation of the variation predictable from the noisy input.

For a centered clean patch $\vec p$ and independent noise $\vec n\sim\mathcal N(0,\mI)$, define the noisy patch $\vec p_t=t\vec p+\sigma\vec n$ with $\sigma=1-t$ and patch velocity $\vec v_p=\vec p-\vec n$. Write $\bm\Sigma_{tt}=\Cov(\vec p_t)$ and $\bm\Sigma_{vt}=\Cov(\vec v_p,\vec p_t)$. Their minimum-MSE affine map satisfies
\begin{equation}
 \mA_v=\bm\Sigma_{vt}\bm\Sigma_{tt}^{-1},\qquad
 \Cov(\widehat{\vec v}_p)=\bm\Sigma_{vt}\bm\Sigma_{tt}^{-1}\bm\Sigma_{tv},\qquad \widehat{\vec v}_p=\mA_v\vec p_t.
\end{equation}

For fixed $t<1$, $\sigma=1-t>0$. Independence of clean patches and unit Gaussian noise gives
\begin{equation}
\bm\Sigma_{tt}=t^2\bm\Sigma_p+\sigma^2\mI,\qquad
\bm\Sigma_{vt}=t\bm\Sigma_p-\sigma\mI.
\end{equation}
In the clean covariance eigenbasis, the covariance of the optimal affine velocity prediction therefore has entries
\begin{equation}
q_i^v(t)=\frac{(t\lambda_i-\sigma)^2}{t^2\lambda_i+\sigma^2}.
\label{eq:affine-velocity-spectrum}
\end{equation}
The corresponding affine clean prediction has covariance entries
\begin{equation}
q_i^p(t)=\frac{t^2\lambda_i^2}{t^2\lambda_i+\sigma^2}.\label{eq:affine-clean-spectrum}
\end{equation}
For fixed $t<1$, these quantities approach 1 and 0, respectively, as $\lambda_i$ approaches zero. A weak clean direction can thus contribute predictably to velocity even when it contributes almost nothing to the clean estimate. These are optimal affine predictors determined by the second moments. Table~\ref{tab:affine-ranks} uses the same ImageNet $16\times16$ patch spectrum for both targets, sorting each derived spectrum independently before computing $r_{90}$.

\begin{table}[!htbp]
\centering\small
\caption{\textbf{Affine clean prediction remains concentrated across noise levels.} Each entry gives the number of directions explaining $90\%$ of predicted variance. Compare clean and velocity at each $t$.}
\label{tab:affine-ranks}
\begin{tabular}{@{}lrrrrrr@{}}
\toprule
Paper $t$ & 0.1 & 0.3 & 0.5 & 0.7 & 0.75 & 0.9 \\
\midrule
Affine clean $r_{90}$ & 1 & 2 & 3 & 5 & 5 & 7 \\
Affine velocity $r_{90}$ & 674 & 656 & 638 & 603 & 588 & 488 \\
\bottomrule
\end{tabular}
\end{table}

The corresponding affine coefficients are $a_i^x=t\lambda_i/(t^2\lambda_i+\sigma^2)$ and $a_i^v=(t\lambda_i-\sigma)/(t^2\lambda_i+\sigma^2)$. For fixed $t<1$, as $\lambda_i\to0$, these coefficients tend to zero and $-1/\sigma$, respectively. The learned embedding gain is measured separately as described in Appendix~\ref{app:embedding-metrics}.

\begin{figure}[!htbp]
\centering\includegraphics[width=.85\linewidth]{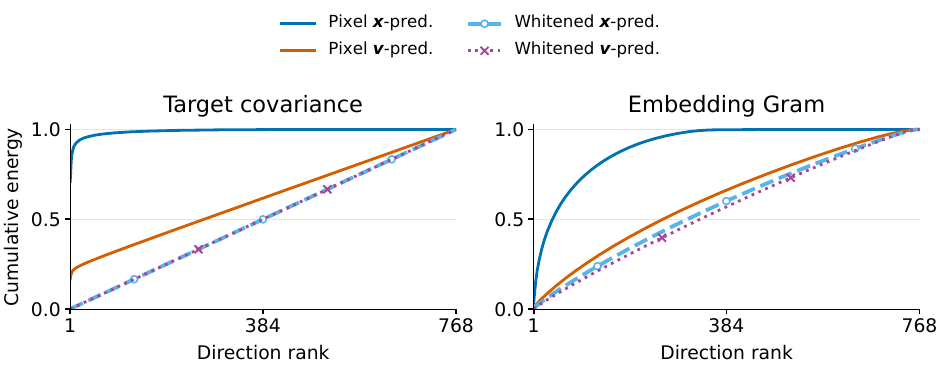}
\caption{\textbf{Target and embedding spectra across coordinates.} Cumulative target covariance (left) and embedding-Gram eigenvalues (right), each sorted independently and normalized by its sum. Pixel $\vec x$-prediction has the most concentrated target and embedding spectra; whitening removes this concentration.}
\label{fig:prediction-covariance-gram}
\end{figure}

For the whitened covariance, $\lambda_i=1$ in every direction, giving affine clean-prediction variance $t^2/(t^2+(1-t)^2)$ in each coordinate. For comparison with these affine predictions, Figure~\ref{fig:prediction-covariance-gram} reports the raw target and learned embedding-Gram spectra.

\subsection{Embedding metrics}
\label{app:embedding-metrics}
For a clean-patch principal direction $\vec{u}_i$, the embedding's squared directional gain is
\begin{equation}
g_i=\left\|
\mW_{\mathrm{in}}\vec{u}_i
\right\|_2^2.
\label{eq:directional-gain}
\end{equation}

We normalize clean energy and embedding gain across these directions as
\begin{equation}
p_i=
\frac{\lambda_i}{\sum_j\lambda_j},
\qquad
\bar g_i
=
\frac{g_i}{\sum_jg_j}.
\label{eq:normalized-energy-gain}
\end{equation}
The normalized gain $\bar g_i$ measures the fraction of total embedding gain assigned to direction $i$, while $p_i$ gives that direction's fraction of clean-patch variance.

The input-side Gram matrix $\mG=\mW_{\mathrm{in}}^\top\mW_{\mathrm{in}}$ gives the same directional gain as $g_i=\vec u_i^\top\mG\vec u_i$, and the top-$k$ gain share is $\sum_{i\le k}\bar g_i$. Subspace alignment is $\|\mU_k^\top\mV_k\|_F^2/k$, where $\mU_k$ and $\mV_k$ contain the leading $k$ eigenvectors of the clean covariance and embedding Gram matrix, respectively.

\begin{table}[!htbp]
\centering\small\setlength{\tabcolsep}{5pt}
\caption{Embedding gain assigned to the leading and trailing 256 clean-PCA directions, as a percentage of total gain over all 768 directions. Pixel $\vec x$ is the plain DiT with $\vec x$-prediction, and all other models use $\vec v$-prediction.}
\label{tab:gain-shares}
\begin{tabular}{@{}lrrrrrr@{}}
\toprule
Gain share (\%) & Pixel, $\vec v$ & Pixel, $\vec x$ & DeCo & PixelDiT & DiP & HyperDiT \\
\midrule
Top 256 & 41.7 & 95.3 & 96.9 & 98.0 & 86.5 & 97.9 \\
Bottom 256 & 28.0 & 0.2 & 0.2 & 0.3 & 5.4 & 0.3 \\
\bottomrule
\end{tabular}
\end{table}

\begin{table}[!htbp]
\centering\scriptsize
\caption{\textbf{Embeddings under $\vec x$-prediction and decoupled designs align with the leading clean-patch subspace.} Higher mean squared cosine indicates stronger overlap. Columns expand the compared subspaces from $k=16$ to $k=256$.}
\begin{tabular}{@{}lrrrrr@{}}
\toprule
Model & $k=16$ & $k=32$ & $k=64$ & $k=128$ & $k=256$ \\
\midrule
Pixel, $\vec x$ & 0.903 & 0.925 & 0.921 & 0.909 & 0.949 \\
Pixel, $\vec v$ & 0.076 & 0.095 & 0.268 & 0.530 & 0.652 \\
DeCo-B & 0.948 & 0.946 & 0.930 & 0.927 & 0.929 \\
PixelDiT-B & 0.920 & 0.877 & 0.885 & 0.891 & 0.935 \\
DiP-B & 0.924 & 0.929 & 0.884 & 0.726 & 0.770 \\
HyperDiT-B & 0.925 & 0.867 & 0.874 & 0.891 & 0.912 \\
\bottomrule
\end{tabular}
\end{table}


\subsection{Target spectra in DINO space}
\label{app:dino-spectrum}
In DINO space, the clean target is the normalized DINOv2-B latent used by RAE, one 768-dimensional feature per token. We estimate its covariance from 100,000 class-balanced ImageNet training images, using all 256 spatial tokens per image (25.6 million observations). Features use the encoder's final LayerNorm followed by RAE's released position-dependent normalization, without additional per-token RMS normalization. The covariance is centered globally over the pooled tokens.

\begin{figure}[!htbp]
\centering
\includegraphics[width=\linewidth]{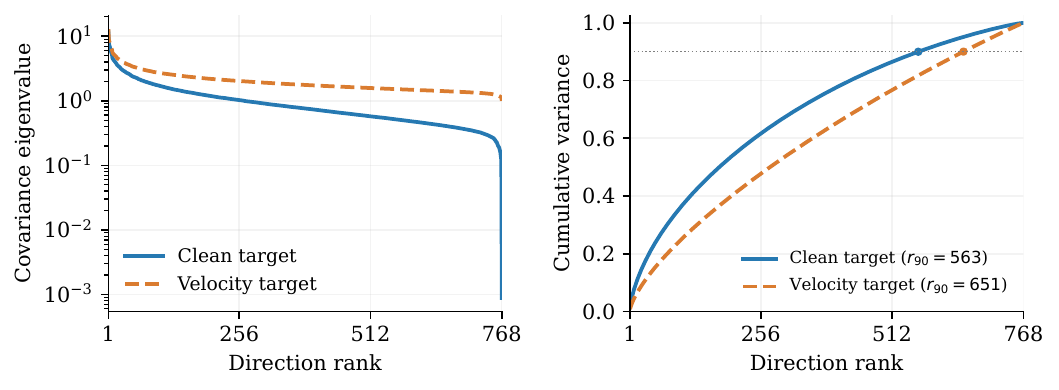}
\caption{\textbf{DINO targets are broad under both prediction choices.} Left: covariance eigenvalues of RAE-normalized DINOv2-B tokens and the corresponding velocity target. Right: cumulative variance, with markers at $r_{90}$. Adding unit Gaussian variance broadens an already distributed clean spectrum.}
\label{fig:dino-target-spectra}
\end{figure}

Writing $\bm\Sigma_{\mathrm D}$ for this measured clean covariance, independent unit-variance Gaussian corruption gives
\begin{equation}
\operatorname{Cov}(\vec v)=\bm\Sigma_{\mathrm D}+\mI,
\qquad \lambda_i^{\vec v}=\lambda_i^{\vec x}+1.
\label{eq:dino-target-covariance}
\end{equation}
Figure~\ref{fig:dino-target-spectra} plots the measured clean spectrum and the velocity spectrum derived from this identity. The clean DINO target already spreads its variance across many directions: $r_{90}=563$ and $\er=579.6$. The velocity target is broader still, with $r_{90}=651$ and $\er=708.3$ (Table~\ref{tab:dino-target-ranks}). Unlike pixel patches, whose $r_{90}$ rises from 8 to 668, DINO features have broad clean variation before corruption is added. DINO and pixel velocity targets thus have similar $r_{90}$, while their final-boundary linear-probe accuracies differ markedly (Appendix~\ref{app:probes}).

\begin{table}[!htbp]
\centering\small
\caption{\textbf{DINO target ranks.} Statistics use the 768-dimensional spectra in Figure~\ref{fig:dino-target-spectra}; definitions follow Appendix~\ref{app:pca-data}.}
\label{tab:dino-target-ranks}
\begin{tabular}{@{}lrrrrr@{}}
\toprule
Target & ER & SR & $r_{90}$ & $r_{95}$ & $r_{99}$ \\
\midrule
DINO, $\vec x$ & 579.6 & 66.84 & 563 & 649 & 735 \\
DINO, $\vec v$ & 708.3 & 121.31 & 651 & 708 & 755 \\
\bottomrule
\end{tabular}
\end{table}

\subsection{Embedding spectra and main-stream output interfaces}
\label{app:endpoints}

\begin{figure}[!htbp]
\centering\includegraphics[width=\linewidth]{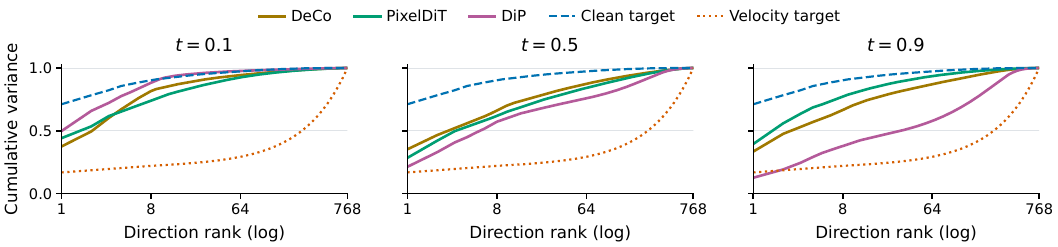}
\par\vspace{-5pt}
\caption{\textbf{Semantic endpoints concentrate variance relative to the velocity target.} Panels fix $t=0.1,0.5,0.9$. Solid curves show the three semantic endpoints, with clean and velocity target spectra as dashed and dotted references. Direction rank uses a logarithmic axis to make the leading subspace visible.}
\label{fig:endpoint-spectra}
\end{figure}
\begin{figure}[!htbp]
\centering\includegraphics[width=\linewidth]{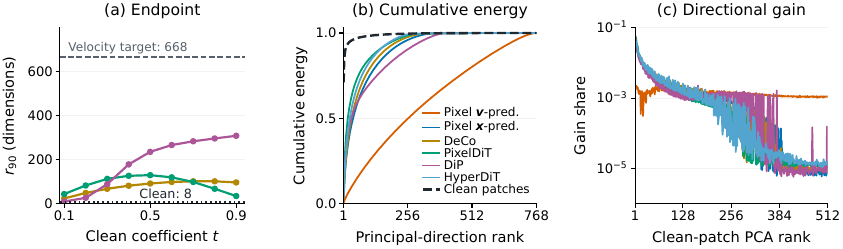}
\par\vspace{-5pt}
\caption{\textbf{Decoupled $\vec v$-prediction models concentrate both main-stream interfaces.} (a) Lower $r_{90}$ means fewer directions explain $90\%$ of output-feature variance. The dashed lines mark raw-target references. (b) Faster-rising curves indicate a more concentrated embedding spectrum. (c) Gain along clean-PCA directions reveals which input variations enter the stream: pixel $\vec v$-prediction retains a broader range than $\vec x$-prediction and the decoupled models.}
\label{fig:spectrum-gain-endpoint}\label{fig:semantic-endpoints-supp}
\end{figure}

Figure~\ref{fig:spectrum-gain-endpoint} reports embedding spectra, directional gains and main-stream output spectra. For decoupled models, the output measurement is taken at the interface that conditions the pixel decoder.

Endpoint measurements use raw epoch-200 B-size checkpoints and the registered 100k ImageNet training images, with true-class conditioning and one deterministic Gaussian draw per image shared across models and times. We record the 768-dimensional features passed from the main Transformer to the detail module, excluding decoder activations, at $t=0.1,0.2,\ldots,0.9$. At $t=0.9$, $r_{90}$ is 97 for DeCo-B, 35 for PixelDiT-B, and 309 for DiP-B. The raw velocity target has covariance $\bm{\Sigma}_p+\mI$ and a time-independent $r_{90}=668$, whereas noisy input patches have covariance $t^2\bm{\Sigma}_p+(1-t)^2\mI$ and a time-dependent spectrum. Figures~\ref{fig:semantic-endpoints} and~\ref{fig:endpoint-spectra} use target spectra as references, with the latter selecting three noise levels and a logarithmic rank axis.

\paragraph{Raw velocity-target baseline.}
The plotted reference is the covariance of the raw patch target $\vec v_p=\vec p-\vec n$. Independence gives $\Cov(\vec v_p)=\bm{\Sigma}_p+\mI$, so its eigenvalues are $\lambda_i+1$. The leading 668 directions explain $90\%$ of raw-target variance, giving $r_{90}=668$.

\begin{table}[!htbp]
\centering\small
\caption{\textbf{Patch covariance and endpoint measurement protocols.} Each covariance treats image-position pairs as observations.}
\label{tab:geometry-protocol}
\begin{tabularx}{\linewidth}{@{}lX@{}}
\toprule
Setting & Value \\
\midrule
Covariance fitting set & 100,000 ImageNet training images, 100 per class \\
Selection / flip seed & 20260806 / 20260807 \\
Preprocessing & ADM center crop to $256^2$, RGB in $[-1,1]$; one deterministic horizontal-flip draw per selected image \\
Patch coordinates & Non-overlapping $16\times16$ RGB; channel, patch row, patch column order \\
Observations & $100{,}000\times256=25.6$M patches of dimension 768 \\
Centering & One global 768-dimensional mean over all patches \\
Numerics & FP32 batch cross-products, FP64 merging and eigendecomposition; TF32 disabled \\
Whitening & All 768 PCA coordinates, divided by their fitted standard deviations \\
Endpoint weights & Raw epoch 200, step 250,200 \\
Endpoint conditioning & True ImageNet class; one shared deterministic noise draw per image \\
Endpoint times / features & $t=0.1,0.2,\ldots,0.9$; 768-dimensional main-stream output to the detail module \\
Embedding measurement & Eigenvalues/eigenvectors of $\bm W_{\mathrm{in}}^\top\bm W_{\mathrm{in}}$ from raw weights \\
\bottomrule
\end{tabularx}
\end{table}

\subsection{A learned long skip for velocity prediction}
\label{app:learned-long-skip}
A learned scalar long skip provides a simple instance of decoupled prediction. We augment JiT-B/16 with a direct noisy-input path,
\begin{equation}
\widehat{\vec v}=\alpha(t)\vec x_{t}+\beta(t)f_\theta(\vec x_{t},t,y),
\label{eq:learned-long-skip}
\end{equation}
with $\vec x_t=t\vec x+(1-t)\vec\epsilon$, as in the main text. A two-layer time-conditioned MLP produces the two scalar coefficients, initialized to $\alpha=0$ and $\beta=1$. The backbone uses a full-rank patch embedding, without context tokens or REPA, and the training objective remains $\vec v$-prediction.

\begin{figure}[!htbp]
\centering
\includegraphics[width=\linewidth]{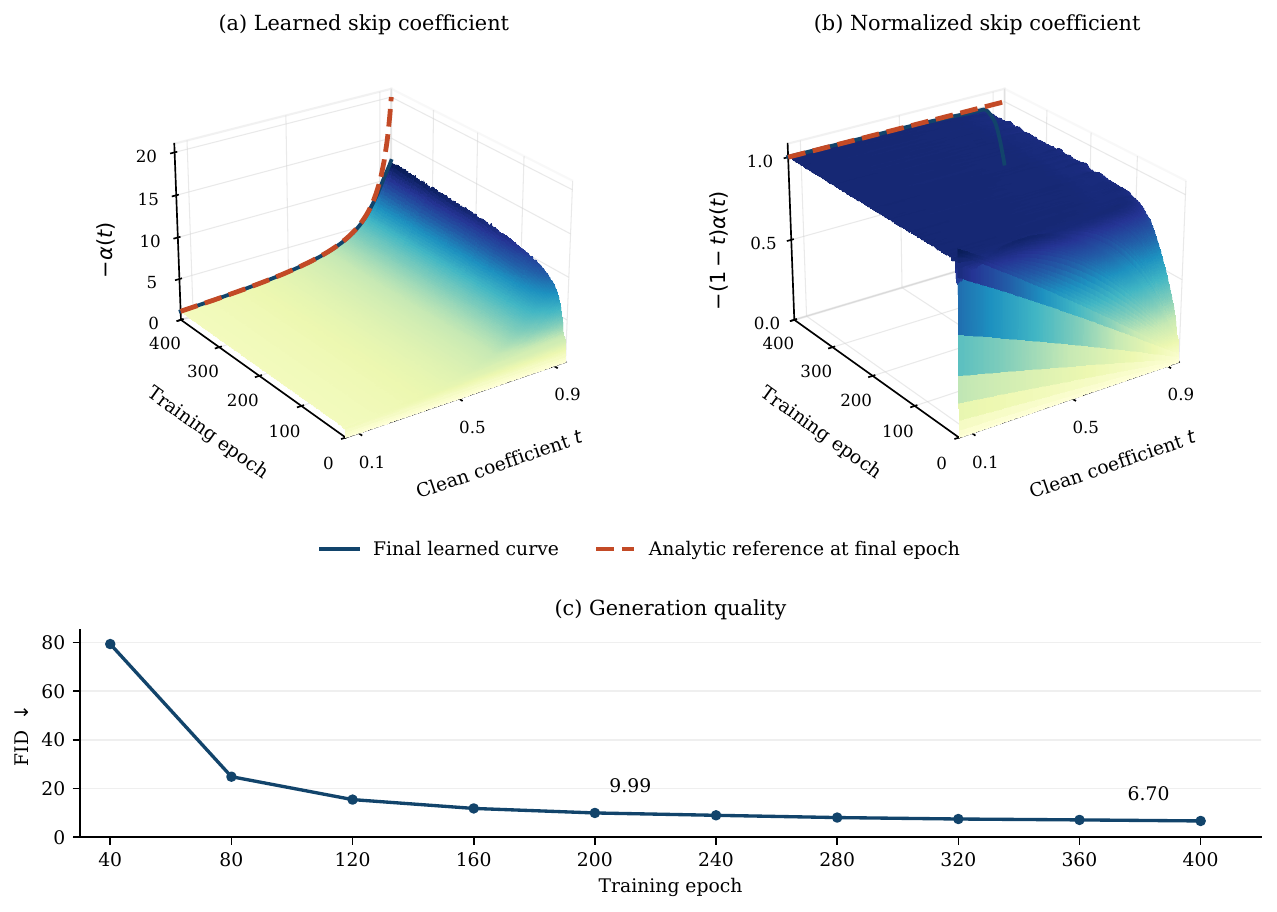}
\caption{\textbf{The learned input skip approaches the clean-to-velocity coefficient during training.} Both surfaces show the clean coefficient $t$, training epoch and the learned skip coefficient. (a) $-\alpha(t)$, with the analytic reference $1/(1-t)$ drawn at the final recorded epoch. (b) $-(1-t)\alpha(t)$, with the corresponding reference at one. Solid blue lines show the final learned curves; dashed orange lines show the references. The surfaces use 246 recorded training snapshots, with denser sampling early in training and shape-preserving interpolation between the 20 measured time points. (c) FID at all ten evaluated checkpoints, using 50,000 samples, EMA weights and Heun-50 sampling; markers denote measurements.}
\label{fig:learned-long-skip}
\end{figure}

Figure~\ref{fig:learned-long-skip} shows that the normalized coefficient $-(1-t)\alpha(t)$ approaches one at low and intermediate $t$ within two epochs and across most of the time grid by epoch 40. Thus the learned input coefficient approaches $-1/(1-t)$, the input coefficient of the clean-to-velocity conversion $\widehat{\vec v}=(\widehat{\vec x}-\vec x_t)/(1-t)$. At 200 epochs, FID is $9.99$ compared with $139.83$ for the plain $\vec v$-prediction backbone under the same recipe, and continued training reaches $6.70$ at 400 epochs.

For comparison, $k$-Diff reports that its learned prediction parameter moves toward clean prediction in pixel space \citep{jin2026kdiff}. In our experiment, the skip starts at zero and the velocity loss acts on the sum of two independently weighted branches.

\subsection{Invertible patch whitening}
\label{app:whitening}
Whitening tests the role of patch spectral concentration by flattening clean-patch variance while preserving manifold structure (Section~\ref{sec:whitening}). An invertible linear map takes a data manifold to one of the same intrinsic dimension; training the same model in this space raises clean-prediction FID from $10.19$ to $132.95$.

Because a DiT receives images through patch embeddings, we apply whitening to each patch, following the emphasis on architecture-dependent input geometry in \citet{liang2026sparseconnectivity}. For centered patches $\vec p$, PCA rows $\bm B$ and scales $\bm D=\operatorname{diag}(\sqrt{\lambda_i})$ define $\vec z=\bm D^{-1}\bm B\vec p$ (Table~\ref{tab:geometry-protocol}). Applied to every patch, this is an invertible linear map of the whole image, and it sets every eigenvalue of the fitted clean-patch covariance to one. Flow matching then runs in this space as it does in pixel space, and generated samples are mapped back by $\bm B^\top\bm D\vec z+\vec\mu$ before FID evaluation.

Both targets use the direct JiT-B/16 backbone, the training settings in Table~\ref{tab:pixel-training} and the sampling protocol in Table~\ref{tab:sampling-protocol}. The whitened velocity target has effective rank $768$, stable rank $768$ and $90\%$ variance rank $692$, and reaches FID $212.54$. Table~\ref{tab:whitening-history} gives the evaluation history.

\begin{table}[!htbp]
\centering\small
\caption{\textbf{Training history in whitened space.} Lower FID and higher IS are better.}
\label{tab:whitening-history}
\begin{tabular}{@{}rrrrr@{}}\toprule
Epoch & Clean FID & Clean IS & Velocity FID & Velocity IS\\\midrule
40 & 298.46 & 2.60 & 238.82 & 2.85\\
80 & 166.89 & 5.82 & 243.77 & 2.29\\
120 & 148.95 & 7.64 & 204.79 & 2.55\\
160 & 140.39 & 8.61 & 209.37 & 3.01\\
200 & 132.95 & 9.54 & 212.54 & 3.35\\
\bottomrule\end{tabular}
\end{table}

Embedding measurements use raw epoch-200 weights, flattened to $768\times768$ in channel--row--column order, with FP64 singular-value decomposition. For singular values $s_i$, Gram energy is $s_i^2$, effective rank is $\exp(-\sum_i p_i\log p_i)$ with $p_i=s_i^2/\sum_j s_j^2$, and $r_{90}$ counts directions containing $90\%$ of that energy. In native whitened coordinates, clean/velocity Gram effective ranks are $738.07/747.69$ and $r_{90}$ values are $647/656$. The raw-pixel clean/velocity controls have effective ranks $175.06/696.24$ and $r_{90}$ values $192/616$. Figure~\ref{fig:prediction-covariance-gram} compares native-coordinate weight spectra, and Figure~\ref{fig:target-embedding} shows alignment and directional gain. Its heatmaps show squared overlaps between the leading 64 input-reference directions and the leading 64 eigenvectors of $\bm W_{\mathrm{in}}^\top\bm W_{\mathrm{in}}$. Raw-pixel references are clean-patch covariance eigenvectors. In whitened coordinates, we retain the coordinate axes defined by the rows of $\bm B$, ordered by the original patch variance, so the reference basis is $\mI$.

\FloatBarrier
\section{Normalized Representation Sensitivity}
\label{app:sensitivity}

We measure how a normalized representation changes when the same clean image receives a new noise draw. Table~\ref{tab:sensitivity-protocol} specifies the shared protocol for pixel, whitened and decoupled models.

\subsection{Observation points and protocol}
\begin{table}[!htbp]
\centering\small
\caption{\textbf{Normalized sensitivity protocol.} The spatial token grid is preserved when computing variance.}
\label{tab:sensitivity-protocol}
\begin{tabularx}{\linewidth}{@{}lX@{}}
\toprule
Setting & Value \\
\midrule
Weights & Raw epoch 200 / step 250,200 \\
Images & 2,048 ImageNet training images, uniform sampling without replacement \\
Corruption & 128 Gaussian draws per image at each $t\in\{0.25,0.50,0.75\}$ \\
Pairing & Identical image indices and deterministic Gaussian draws across models/times \\
Input / conditioning & Pixel or whitened training coordinates (Appendix~\ref{app:whitening}); null class 1000; no image flips \\
Observation & Attention/MLP input before normalization and adaptive modulation \\
Readout & Per-token RMS, $\epsilon=10^{-6}$; keep all 256 spatial positions \\
Observed width & 768; $D=256\times768=196{,}608$ coordinates \\
Depth & 12 main-stream blocks; 8 for HyperDiT \\
Token selection & Spatial main-stream tokens; context and register tokens omitted \\
Numerics & BF16 forward, FP32 normalization, FP64 variance accumulation \\

\bottomrule
\end{tabularx}
\end{table}

\begin{figure}[!htbp]
\centering\includegraphics[width=\linewidth]{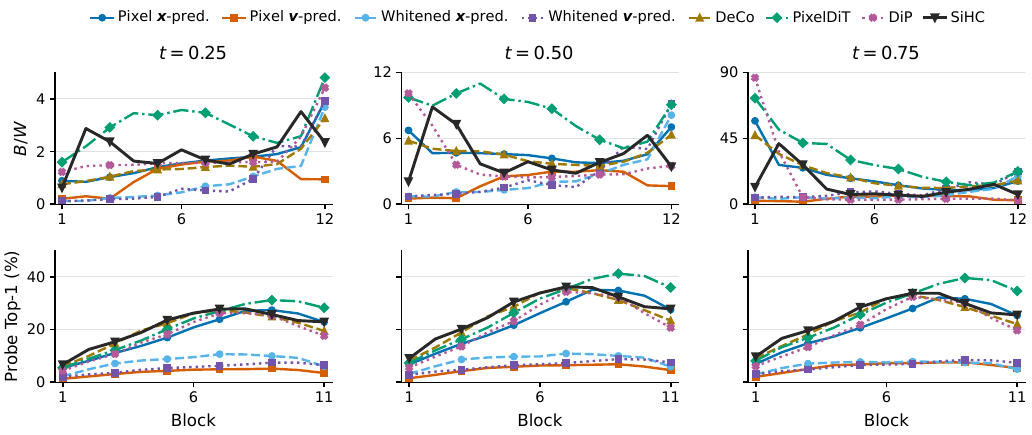}
\par\vspace{-5pt}
\caption{\textbf{Image-to-noise ratio and class readability through depth.} Top: variation between images relative to variation from noise ($B/W$). Bottom: class readability measured by linear-probe Top-1 accuracy. Curves cover the plain DiT in pixel and whitened space, the decoupled models and SiHC. Columns fix the clean coefficient $t$.}
\label{fig:plain-target-diagnostics}
\end{figure}

The top row of Figure~\ref{fig:plain-target-diagnostics} selects the MLP input of each block. Its lower row uses next-attention inputs after blocks 1--11, with the probe fitting and spatial pooling described in Appendix~\ref{app:probes}. All measurements use raw epoch-200 weights and precede native normalization and modulation. SiHC uses the workspace read from its persistent states in both rows. Figure~\ref{fig:sensitivity-full} retains both attention and MLP observations in execution order. SiHC workspace measurements and the main SiHC progression in Table~\ref{tab:progression} use the sublayerwise implementation, whose attention and MLP have separate read/write maps.

\subsection{Normalized noise variation and its image-dependent counterpart}

Let $\vec h(\vec x,\vec\epsilon;t)$ denote the normalized layer input for image $\vec x$ and noise $\vec\epsilon$, with $\bm\mu(\vec x;t)=\E_{\vec\epsilon}\vec h(\vec x,\vec\epsilon;t)$. Variation between images and variation from resampled noise are
\begin{equation}
B(t)=\E_{\vec x}\|\bm\mu(\vec x;t)-\E_{\vec x'}\bm\mu(\vec x';t)\|_2^2,\qquad
W(t)=\E_{\vec x,\vec\epsilon}\|\vec h(\vec x,\vec\epsilon;t)-\bm\mu(\vec x;t)\|_2^2.
\label{eq:between-within}
\end{equation}
The representation signal-to-noise ratio $B/W$ measures between-image variation relative to corruption variation. The signal here consists of all image-dependent differences.

Let $\vec r_{ijk}\in\mathbb R^C$ be spatial token $k$ for image $i$ and noise draw $j$ at a fixed observation point and time. We apply a separate, non-affine RMS normalization,
\begin{equation}
\widetilde{\vec r}_{ijk}
=\frac{\vec r_{ijk}}{\sqrt{\|\vec r_{ijk}\|_2^2/C+10^{-6}}},
\end{equation}
then concatenate the spatial tokens conceptually into $\vec h_{ij}\in\mathbb R^D$. Spatial locations remain separate throughout the variance computation. For $N=2048$ images and $K=128$ noises, define the image means $\bar{\vec h}_i=K^{-1}\sum_j\vec h_{ij}$ and their grand mean $\bar{\vec h}=N^{-1}\sum_i\bar{\vec h}_i$. The within-image covariance trace and the plotted sensitivity are
\begin{equation}
\widehat W=\frac{1}{N(K-1)}\sum_{i=1}^{N}\sum_{j=1}^{K}
\|\vec h_{ij}-\bar{\vec h}_i\|_2^2,
\qquad
\widehat S=\frac{\widehat W}{D}.
\label{eq:normalized-sensitivity}
\end{equation}
Thus $S$ is the mean coordinate variance induced by changing the noise while holding the image fixed. At the population level, it is also half the mean squared distance per coordinate between two independently corrupted versions of the same image. Larger $S$ means greater noise sensitivity of the normalized representation.

Small within-image variation is useful only if the representation still distinguishes images. We therefore retain the between-image covariance trace and its finite-$K$ correction,
\begin{align}
\widehat B_{\rm raw}
&=\frac{1}{N-1}\sum_{i=1}^{N}\|\bar{\vec h}_i-\bar{\vec h}\|_2^2,\\
\widehat B&=\widehat B_{\rm raw}-\frac{\widehat W}{K},
\qquad
\widehat R=\frac{\widehat B}{\widehat W}.
\end{align}
The correction removes noise variation left in the finite-sample image means, isolating image-dependent variation in the representation. $B/D$ and $R$ accompany $S$ in Figure~\ref{fig:sensitivity-full}, allowing low sensitivity to be distinguished from a representation that varies little with either images or noise.

\subsection{What changes through depth}

\begin{figure}[!htbp]
\centering\includegraphics[width=\linewidth]{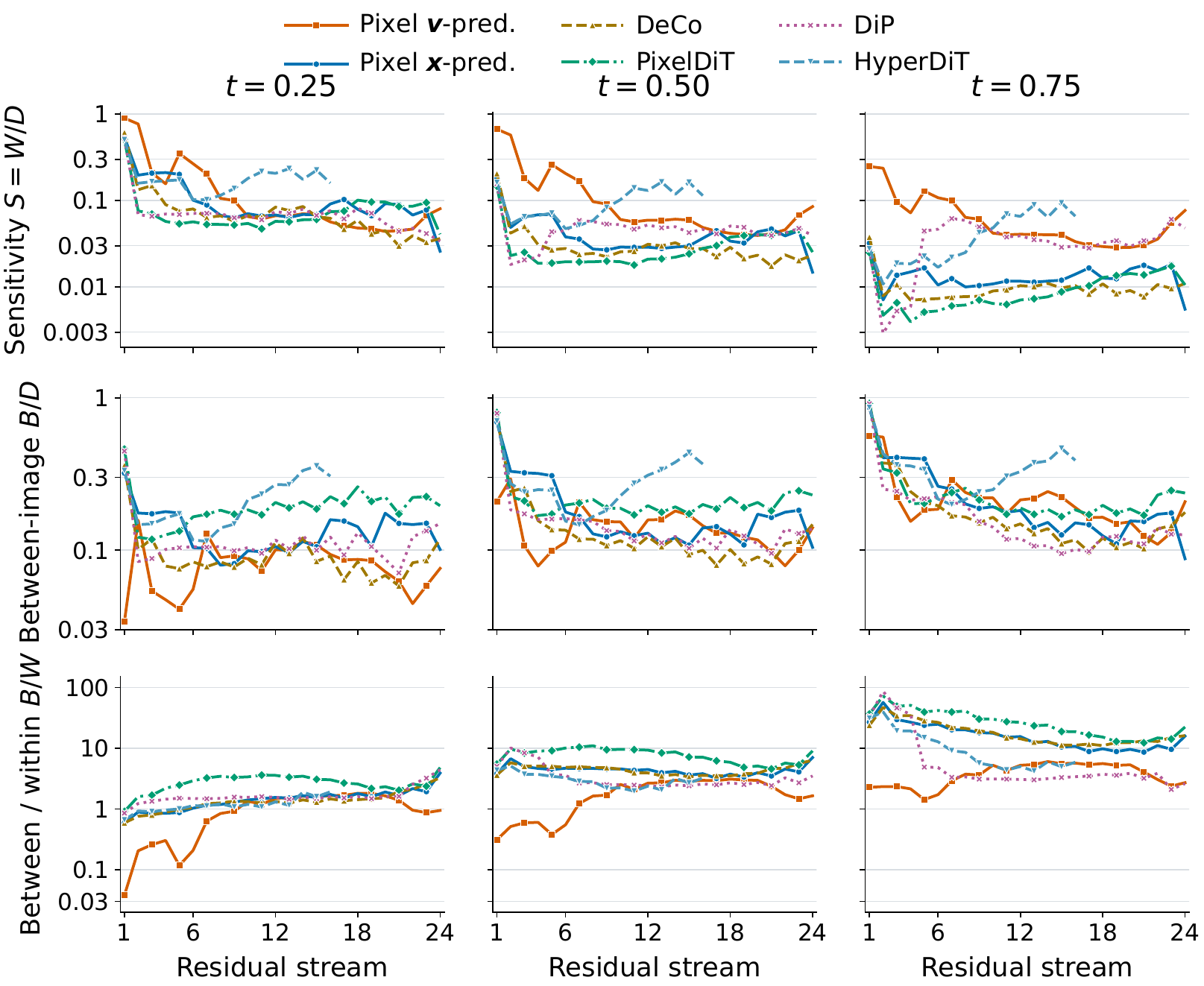}
\par\vspace{-5pt}
\caption{\textbf{Image and noise variation evolve differently through depth.} Columns fix the clean coefficient $t$. The top and middle rows show noise variance $W/D$ and between-image variance $B/D$, with $D$ the state dimension. The bottom row shows their ratio $B/W$. Larger ratios indicate stronger image differences relative to corruption. Vertical axes are logarithmic, with a shared range within each row.}
\label{fig:sensitivity-full}
\end{figure}

Pixel $\vec v$-prediction reduces noise sensitivity through much of the backbone, followed by a late increase at the MLP inputs. At $t=0.50$, $S$ rises from $0.03948$ at block 10 to $0.08516$ at block 12, while pixel $\vec x$-prediction falls from $0.04348$ to $0.01484$. The velocity rebound occurs at all three measured times. Its noise variation grows faster than its between-image variation over these final blocks, so $B/W$ falls (Figure~\ref{fig:sensitivity-full}).

The decoupled models show different depth profiles. DeCo and PixelDiT have lower final MLP-input sensitivity than pixel $\vec v$-prediction at every measured time. PixelDiT also shows a clear decrease at its last attention update. DiP is less sensitive than pixel $\vec v$-prediction at the final measured point but rises late at $t=0.75$, while HyperDiT retains appreciable late sensitivity in its shorter semantic stream. Table~\ref{tab:sensitivity-summary} summarizes the depth averages.

\begin{table}[!htbp]
\centering\scriptsize\setlength{\tabcolsep}{3.5pt}
\caption{\textbf{Depth-averaged image and noise variation.} Each entry averages a per-observation statistic. In particular, $\bar R$ averages the individual ratios rather than dividing the averaged variances. Complete curves appear in Figure~\ref{fig:sensitivity-full}.}
\label{tab:sensitivity-summary}
\begin{tabular}{@{}lrrr|rrr|rrr@{}}
\toprule
& \multicolumn{3}{c}{$t=0.25$} & \multicolumn{3}{c}{$t=0.50$} & \multicolumn{3}{c}{$t=0.75$} \\
Model & $\bar S$ & $\overline{B/D}$ & $\bar R$ & $\bar S$ & $\overline{B/D}$ & $\bar R$ & $\bar S$ & $\overline{B/D}$ & $\bar R$ \\
\midrule
Pixel, $\vec v$ & 0.164 & 0.081 & 1.04 & 0.133 & 0.142 & 1.89 & 0.071 & 0.221 & 3.90 \\
Pixel, $\vec x$ & 0.116 & 0.137 & 1.50 & 0.045 & 0.192 & 4.39 & 0.014 & 0.234 & 17.69 \\
DeCo & 0.090 & 0.098 & 1.41 & 0.035 & 0.149 & 4.44 & 0.010 & 0.196 & 19.07 \\
PixelDiT & 0.086 & 0.197 & 2.78 & 0.032 & 0.224 & 7.87 & 0.010 & 0.241 & 28.83 \\
DiP & 0.083 & 0.121 & 1.72 & 0.049 & 0.162 & 3.56 & 0.036 & 0.182 & 11.24 \\
HyperDiT & 0.189 & 0.216 & 1.19 & 0.099 & 0.293 & 3.07 & 0.044 & 0.357 & 12.25 \\
\bottomrule
\end{tabular}
\end{table}

Whitened models have high late-layer $B/W$ ratios while their linear-probe accuracies remain low (Figure~\ref{fig:plain-target-diagnostics}). Here $B$ measures all image-dependent variation, whereas the probe measures class readability.

\subsection{SiHC workspace sensitivity}
\label{app:sihc-sensitivity}

The $4\times4$ SiHC model follows Table~\ref{tab:sensitivity-protocol}. We observe its read workspace immediately before each attention or MLP update, prior to native normalization and modulation, and use the corresponding inputs for the plain DiT. Both have $D=256\times768$ workspace coordinates.

\begin{figure}[!htbp]
\centering\includegraphics[width=\linewidth]{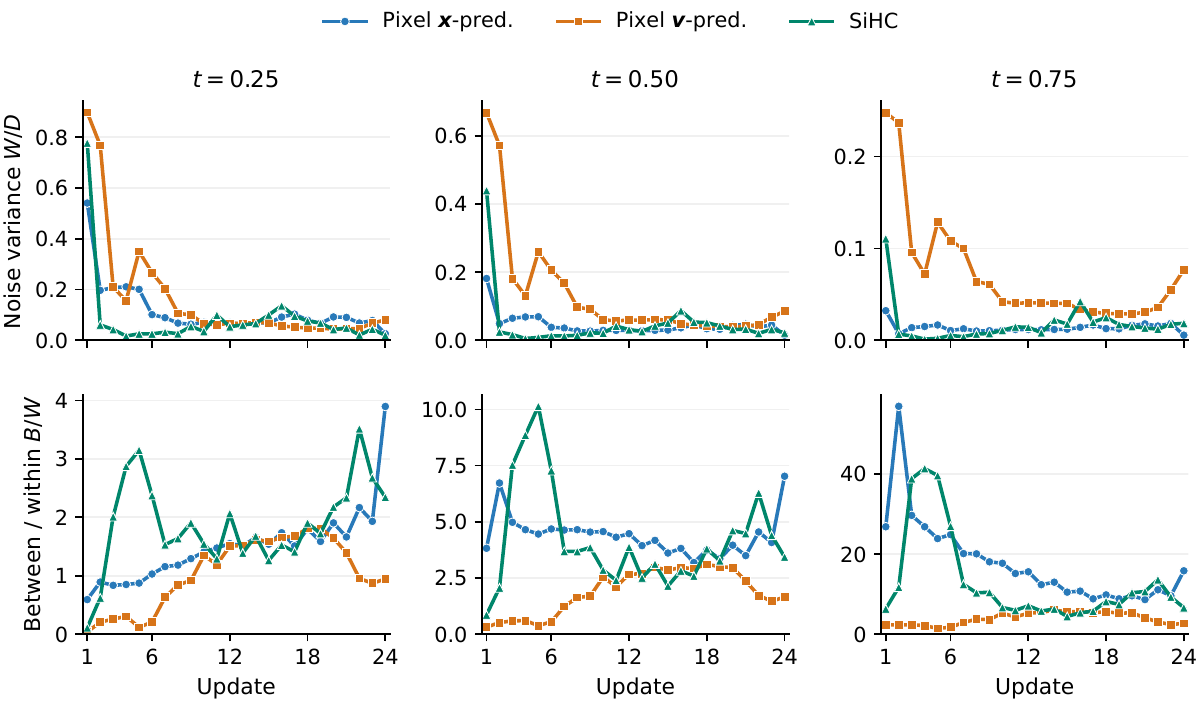}
\par\vspace{-5pt}
\caption{\textbf{SiHC's late workspace is less sensitive to corruption than pixel $\vec v$-prediction.} Columns fix $t$. Top: noise-induced variance per coordinate ($W/D$). Bottom: between-image variation relative to noise ($B/W$). Vertical ranges differ between panels.}
\label{fig:sihc-sensitivity-all}
\end{figure}

The next-layer linear probes in Appendix~\ref{app:probes} observe block 2--12 attention inputs, after blocks 1--11. Sensitivity retains all spatial coordinates, whereas the linear probe spatially averages normalized tokens. These measurements share a representation location but test different properties. Figure~\ref{fig:sihc-sensitivity-all} includes all attention and MLP inputs and separates noise variance from its ratio to image variation.

At the final MLP input, SiHC has lower noise variance and higher $B/W$ than pixel $\vec v$-prediction at all three measured times. For example, at $t=0.50$, their $W/D$ values are $0.0219$ and $0.0852$, and their $B/W$ values are $3.43$ and $1.65$, respectively. Both image and noise variation are lower in SiHC at this location, with a larger relative reduction in noise.

\FloatBarrier
\section{Residual States in the Swiss-Roll Toy}
\label{app:toy}\label{sec:toy}
We use the toy setting of \citet{li2026back}, with a planar Swiss roll embedded in 512 dimensions, and train the same fully connected architecture under $\vec x$- and $\vec v$-prediction. Table~\ref{tab:toy-protocol} lists the data, training and sampling settings.

\begin{table}[!htbp]
\centering\small
\caption{\textbf{Swiss-roll training and sampling settings.}}
\label{tab:toy-protocol}
\begin{tabularx}{\linewidth}{@{}lX@{}}
\toprule
Setting & Value \\
\midrule
Dataset / ambient dimension & 8,192 points / 512; dataset seed 42 \\
Construction & Keep coordinates 0 and 2 of the Swiss roll, project with a Gaussian-QR $512\times2$ basis, standardize each ambient coordinate \\
Standardization & Subtract empirical mean, divide by empirical standard deviation plus $10^{-6}$ \\
Training & 200,000 updates, batch 256, AdamW, constant learning rate $10^{-4}$ \\
Optimizer & $\beta=(0.9,0.999)$, $\epsilon=10^{-8}$, weight decay 0, gradient norm clip 1 \\
Time & Sigmoid-normal; implementation noise time $q=1-t$, minimum $q=0.001$ \\
Generation & 2,048 standard-Gaussian initial points; raw final weights \\
Integrator & Euler on 100 evenly spaced time points from $q=0.999$ to $0.001$ (99 updates) \\
Implementation & JAX \\
\bottomrule
\end{tabularx}
\end{table}

\paragraph{Model.}
The fully connected denoiser uses five AdaLN-zero residual ReLU blocks with residual state and MLP hidden width 256.

\paragraph{State measurements.}
The rank measurements use the final 200k checkpoint and all 8,192 saved training examples, with one paired Gaussian draw per example. They record the persistent state before AdaLN at the input of each of the five blocks, so block indices 0--4 in Figure~\ref{fig:toy-residual-rank} correspond to the inputs of blocks 1--5. Original toy time $q$ corresponds to paper time $t=1-q$.

\FloatBarrier
\section{ImageNet Training, Sampling and Model Configurations}
\label{app:sihc-config}

\subsection{Data and training}
We train on ImageNet-1k \citep{deng2009imagenet} at $256\times256$ resolution, with 1,281,167 images and 1,000 classes. We use ADM-style center cropping and random horizontal flips, and scale pixel values to $[-1,1]$. Table~\ref{tab:pixel-training} summarizes the shared pixel-space training settings.

\begin{table}[!htbp]
\centering\small
\caption{\textbf{Pixel-space training settings.} B/L/H scaling uses 200 epochs without representation alignment. The extended H and XL runs use 600 epochs.}
\label{tab:pixel-training}
\begin{tabularx}{\linewidth}{@{}lX@{}}
\toprule
Setting & Value \\
\midrule
Image resolution & $256\times256$ RGB \\
Training duration & 200 epochs (controlled/scaling); 600 (extended H/XL) \\
Optimizer & AdamW, $\beta_1=0.9$, $\beta_2=0.95$ \\
Global batch size & 1,024 \\
Learning rate & $2\times10^{-4}$, constant after warmup \\
Weight decay & 0 \\
Warmup & 5 epochs \\
Time distribution & $\operatorname{logit}(t)\sim\mathcal N(-0.8,0.8^2)$ \\
Gaussian noise scale & 1.0 \\
Division floor & $\max(1-t,0.05)$ \\
Class-conditioning dropout & 0.1 \\
EMA decay & 0.9999 \\
Representation alignment & None for controlled runs and B/L/H scaling; REPA for XL \\
\bottomrule
\end{tabularx}
\end{table}

\paragraph{Prediction and loss.}
For a clean image $\vec x$ and $\vec\epsilon\sim\mathcal N(0,\mI)$, we form $\vec x_t=t\vec x+(1-t)\vec\epsilon$. $\vec v$-prediction models minimize mean squared error to $(\vec x-\vec x_t)/\max(1-t,0.05)$. $\vec x$-predictions are converted to velocity with the same denominator before computing the loss. The division floor stabilizes training near $t=1$.

\paragraph{Optimization.}
Controlled models use the optimizer and learning-rate schedule in Table~\ref{tab:pixel-training} and train for 200 epochs. Generation uses EMA weights, while class-conditioning dropout trains the unconditional branch used by classifier-free guidance.

\paragraph{Model configurations.}
Both $\vec x$- and $\vec v$-prediction JiT-B/16 baselines use a direct linear patch embedding without a bottleneck and omit in-context class tokens. Their backbone has 12 blocks, width 768 and 12 attention heads. The main SiHC-B ablations likewise omit in-context tokens. For scaling, all SiHC sizes use 32 in-context class tokens, with the configurations in Table~\ref{tab:sihc-model-config}.

The scaling configurations embed each $4\times4$ RGB subpatch directly into the hidden dimension and group sixteen local states within each $16\times16$ computational patch. Independent static feature-wise read/write maps surround every attention and MLP sublayer, with identity carry between states. The workspace contains 256 spatial tokens and grows to 288 tokens when the class tokens are inserted. Class tokens follow ordinary residual updates and do not enter the spatial read/write maps.

\begin{table}[!htbp]
\centering\small\setlength{\tabcolsep}{6pt}
\caption{\textbf{Configurations of SiHC models.} Block indices are one-based. Parameter counts and GFLOPs cover the denoiser, excluding the REPA encoder and projector.}
\label{tab:sihc-model-config}
\begin{tabular*}{\linewidth}{@{\extracolsep{\fill}}lcccc@{}}
\toprule
 & SiHC-B & SiHC-L & SiHC-XL & SiHC-H\\
\midrule
Depth & 12 & 24 & 40 & 32\\
Hidden dimension & 768 & 1,024 & 1,024 & 1,280\\
Attention heads & 12 & 16 & 16 & 16\\
First block with context & 5 & 9 & 9 & 9\\
REPA & No & No & Yes & No\\
Parameters (M) & 131.0 & 459.5 & 762.8 & 953.8\\
GFLOPs/forward & 50.9 & 177.0 & 299.1 & 368.2\\
\midrule
Image resolution & \multicolumn{4}{c}{$256\times256$}\\
Computational / local patch & \multicolumn{4}{c}{$16\times16$ / $4\times4$}\\
Streams per computational patch & \multicolumn{4}{c}{16}\\
In-context class tokens & \multicolumn{4}{c}{32}\\
Input embedding & \multicolumn{4}{c}{Direct linear map, no bottleneck}\\
Read/write maps & \multicolumn{4}{c}{Static feature-wise, per sublayer}\\
Carry & \multicolumn{4}{c}{Identity}\\
\bottomrule
\end{tabular*}
\end{table}

\subsection{Sampling and evaluation}
\begin{table}[!htbp]
\centering\small\setlength{\tabcolsep}{4pt}
\caption{\textbf{Generation protocols for our experiments.} All rows use 50,000 samples. Guidance intervals are in the paper's clean-time convention.}
\label{tab:sampling-protocol}
\begin{tabular}{@{}lllll@{}}
\toprule
Experiment & Solver & EMA & CFG & Interval \\
\midrule
Pixel B-size / ablations & Heun50 & 0.9999 & 2.9 & $[0.1,1]$ \\
SiHC-L & Heun50 & 0.9999 & 2.4 & $[0.1,1]$ \\
SiHC-H & Heun50 & 0.9999 & 2.1 & $[0.1,0.9]$ \\
SiHC-XL + REPA & Heun50 & 0.9999 & 2.4 & $[0.1,0.9]$ \\
Whitened $\vec x$ / $\vec v$ & Heun50 & 0.9999 & 2.9 & $[0.1,1]$ \\
DINO/RAE plain / mHC & Heun50 & 0.9999 & 2.9 & $[0.1,1]$ \\
\bottomrule
\end{tabular}
\end{table}

For pixel sampling, the conditional and null-class predictions define the guided velocity $\widehat v_g=\widehat v_\varnothing+w(t)(\widehat v_y-\widehat v_\varnothing)$, where $w(t)$ is the listed CFG scale inside the guidance interval and 1 outside. Standard Gaussian initial states have noise scale one. FID uses 50,000 generated images, balanced across classes, and the JiT ImageNet-$256^2$ reference statistics. Samples are converted to uint8 RGB for evaluation. Whitened outputs are inverse-transformed before conversion to RGB. Diagnostic figures use raw weights; generation uses the EMA state specified in Table~\ref{tab:sampling-protocol}.

\paragraph{B-size decoupled models.}
Table~\ref{tab:bsize-decoupled-fid} reports the generation trajectories of our DeCo, DiP and PixelDiT runs used in the decoupled-model analysis. Each uses a 12-block, width-768 semantic backbone, $\vec v$-prediction and 200 training epochs without REPA. We evaluate EMA checkpoints every 40 epochs with the pixel sampling protocol in Table~\ref{tab:sampling-protocol}. All three improve throughout training, reaching FID 8.07, 13.31 and 6.33, respectively. The corresponding representation diagnostics use raw weights.

\begin{table}[!htbp]
\centering\small
\caption{\textbf{Generation trajectories of B-size decoupled models.} FID on 50,000 ImageNet-$256^2$ samples at the indicated training epochs, using Heun50 and CFG 2.9. All runs use seed 0.}
\label{tab:bsize-decoupled-fid}
\begin{tabular}{@{}lrrrrr@{}}
\toprule
 & \multicolumn{5}{c}{Training epochs}\\
\cmidrule(l){2-6}
Model & 40 & 80 & 120 & 160 & 200\\
\midrule
DeCo-B & 63.84 & 19.42 & 12.64 & 9.54 & 8.07\\
DiP-B & 82.13 & 27.74 & 19.54 & 15.89 & 13.31\\
PixelDiT-B & 67.44 & 14.83 & 9.44 & 7.35 & 6.33\\
\bottomrule
\end{tabular}
\end{table}

\paragraph{DINO/RAE training.}
The shared RAE configuration uses DINOv2-with-registers base, encoder input size 224 and normalized $768\times16\times16$ latents \citep{oquab2023dinov2}. The frozen ViT-XL decoder and latent normalization follow RAE~\citep{zheng2026rae}. The decoder noise parameter is zero. Diffusion training uses 200 epochs, global batch 1,024, AdamW with learning rate $2\times10^{-4}$, betas $(0.9,0.95)$ and zero weight decay. Learning rate is constant with no warmup, EMA decay is 0.9999 and the training seed is 42. The transport is linear, with logit-normal time sampling and the dimension-dependent time shift used by RAE (latent dimension 196,608, reference dimension 4,096). The mHC variant has four streams and class dropout 0.1. Final evaluation draws 50 samples per class with seed 12345, Heun-50 and CFG $2.9$ over $[0.1,1]$, matching the pixel-space guidance interval. RAE's implementation time runs opposite to the paper's $t$.

\subsection{SiHC ablations}\label{app:ablation}
\paragraph{HC intervention.}
Table~\ref{tab:hc-target} uses B-size models trained for 200 epochs. DINOv2-B features have shape $16\times16\times768$, matching the 256 tokens and 768 coordinates per token obtained by patchifying $256\times256\times3$ images into $16\times16$ RGB patches. Each plain/mHC pair shares its training protocol. The mHC intervention uses four copied streams with dynamic read, write and mixing maps, while keeping attention and MLP widths fixed. Both pixel and DINOv2-B models are evaluated with Heun-50 and CFG $2.9$ over $[0.1,1]$.

\paragraph{Design ablations.}
Table~\ref{tab:ablation-settings} specifies the interfaces behind Table~\ref{tab:progression}. The main SiHC progression uses separate read/write maps for attention and MLP, giving 24 connection events in the 12-block B backbone. All models use computational patch size $16\times16$, 200 training epochs and the B-size sampling protocol. Adding 32 in-context class tokens, as in JiT, to the $4\times4$ model lowers FID from $8.07$ to $6.17$. The lower group separately compares direct and factorized embeddings in the earlier blockwise model, with intermediate width 128 for the latter.

\begin{table}[!htbp]
\centering\small\setlength{\tabcolsep}{4pt}
\caption{\textbf{B-size ablation configurations.} All models use $\vec v$-prediction. $b$ is local input side length in pixels, and $S$ is the number of streams per computational patch.}
\label{tab:ablation-settings}
\begin{tabular}{@{}lrrllr@{}}
\toprule
Configuration & $b$ & $S$ & Read/write & Carry & FID \\
\midrule
mHC copied input & 16 & 4 & mHC & mixed & 25.36 \\
mHC spatial input & 8 & 4 & mHC & mixed & 13.25 \\
Static scalar access & 8 & 4 & scalar & identity & 11.50 \\
SiHC, spatial input & 8 & 4 & feature-wise & identity & 10.10 \\
SiHC, direct input & 4 & 16 & feature-wise & identity & 8.07 \\
\quad + 32 in-context class tokens & 4 & 16 & feature-wise & identity & 6.17 \\
\midrule
\multicolumn{6}{@{}l}{\textit{Earlier blockwise embedding comparison}}\\
SiHC, direct input & 4 & 16 & feature-wise & identity & 8.54 \\
Factorized embedding & 4 & 16 & feature-wise & identity & 7.97 \\
\bottomrule
\end{tabular}
\end{table}

\subsection{Scaling without representation alignment}\label{app:sihc-scaling}
\begin{figure}[!htbp]
\centering
\includegraphics[width=\linewidth]{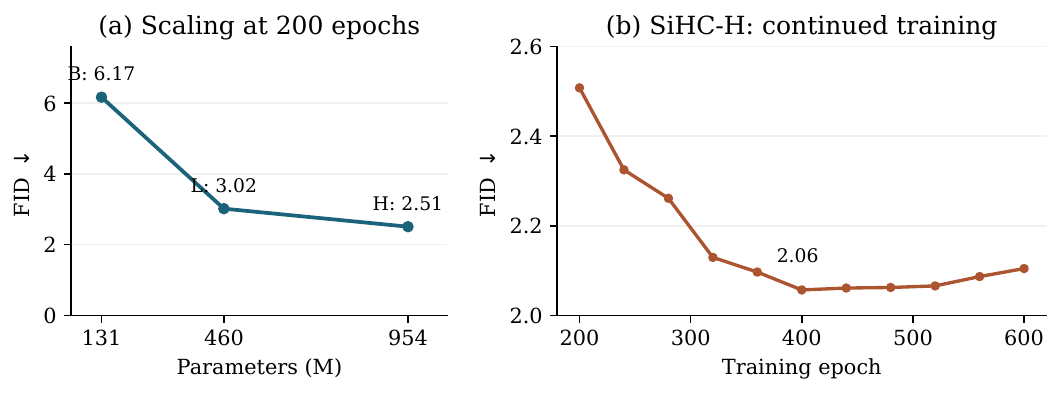}
\caption{\textbf{Scaling SiHC without representation alignment.} Left: B/L/H at 200 epochs, with 32 in-context class tokens and sublayerwise read/write. Right: continued training of SiHC-H, including all measured points from epoch 200 onward. Generation quality levels off around FID $2.06$.}
\label{fig:sihc-scaling}
\end{figure}

SiHC scales effectively under direct $\vec v$-prediction without a pretrained representation encoder. With 32 in-context class tokens and sublayerwise read/write maps, increasing the backbone from B to L and H reduces FID from $6.17$ to $3.02$ and $2.51$ after 200 epochs (Table~\ref{tab:sihc-scaling}). Figure~\ref{fig:sihc-scaling} shows this progression and the subsequent training of SiHC-H, which levels off around FID $2.06$.

\begin{table}[!htbp]
\centering\small\setlength{\tabcolsep}{5pt}
\caption{\textbf{Scaling pixel-space Transformers on ImageNet-$256^2$.} The upper group reproduces DiP's DiT-only results with its Euler-100 protocol \citep{chen2026dip}. The lower group uses 200 training epochs without REPA and Heun-50, with CFG $2.9/2.4/2.1$ for B/L/H. GFLOPs count one denoiser forward with two FLOPs per multiply-add. $^*$Analytical estimates from the published configurations and released backbone (Appendix~\ref{app:compute}).}
\label{tab:sihc-scaling}
\begin{tabular*}{\linewidth}{@{\extracolsep{\fill}}llrrrr@{}}
\toprule
Model & Depth $\times$ width & Params & GFLOPs & FID $\downarrow$ & IS $\uparrow$\\
\midrule
\multicolumn{6}{@{}l}{\textit{DiT-only configurations reported by DiP}}\\
DiT-only & $26\times1152$ & 629M & $221.2^*$ & 5.28 & 243.8\\
DiT-only & $32\times1152$ & 772M & $272.0^*$ & 4.91 & 251.7\\
DiT-only & $26\times1280$ & 776M & $272.0^*$ & 4.28 & 249.6\\
DiT-only & $26\times1536$ & 1.1B & $389.3^*$ & 2.83 & 285.6\\
\midrule
\multicolumn{6}{@{}l}{\textit{SiHC, 200 epochs, without REPA}}\\
SiHC-B & $12\times768$ & 131M & 50.9 & 6.17 & 196.3\\
SiHC-L & $24\times1024$ & 460M & 177.0 & 3.02 & 268.8\\
SiHC-H & $32\times1280$ & 954M & 368.2 & 2.51 & 274.1\\
\bottomrule
\end{tabular*}
\end{table}

DiP's DiT-only study reports FID $2.83$ for its widest 1.1B-parameter model, alongside our SiHC-H result of $2.51$ with 954M parameters under the protocols in Table~\ref{tab:sihc-scaling}.

\subsection{SiHC-XL with representation alignment}\label{app:sihc-repa}
Our XL configuration combines the same sublayerwise read/write interface with REPA \citep{yu2025repa} and reaches FID $1.71$. It uses 40 blocks of width 1024, direct $4\times4$ patch embeddings and 32 in-context class tokens inserted before block 9. Training follows Table~\ref{tab:pixel-training}, with a frozen DINOv3 ViT-L/16 teacher \citep{simeoni2025dinov3}. The teacher receives clean $256\times256$ images with ImageNet normalization. Removing its CLS and four register tokens leaves 256 patch features of dimension 1024.

At block 8, before context tokens enter, we align the sum of the student's MLP read workspace and its gated update. A three-layer projector with dimensions $1024\to2048\to2048\to1024$ and SiLU activations maps these workspace features to the teacher space. The training loss adds $0.5$ times the negative cosine similarity, averaged over patches, to the velocity MSE. The teacher and projector are used only during training. The denoiser contains 762,840,400 parameters and uses 299.10 GFLOPs per forward pass. BF16 autocast and Flash attention are used for training, with the static read/write maps fused as described in Appendix~\ref{app:fusion}.

\paragraph{Reference results and cost accounting.}
Table~\ref{tab:full-performance} reports guided FID on 50,000 samples, with each method's own recipe and training duration. Parameter counts cover the denoiser, excluding pretrained autoencoders and alignment encoders. RAE uses AutoGuidance for its DiT$^{\mathrm{DH}}$ result.

Table~\ref{tab:sihc-model-config} reports parameter counts and forward costs for the listed configurations. FLOPs count dominant matrix operations and leading residual-interface arithmetic, with one multiply-add counted as two FLOPs.

\subsection{Compute accounting}\label{app:compute}
\paragraph{Reference-model compute accounting.}\label{app:reference-compute}
Table~\ref{tab:full-performance} uses the two-FLOPs-per-multiply-add convention for linear, convolution and attention matrix products. DiT, SiT, REPA, LightningDiT and DDT use the published 119-G multiply-add reference, converted to 238 GFLOPs. PixelFlow and PixNerd use 5818 and 268 GFLOPs under the same convention, as reported in \citet{li2026back,chen2026asymflow}. JiT-H/G, DiP-XL, DeCo-XL, PixelDiT-XL and SiHC-XL use the cached architecture counts, rounded to whole GFLOPs. PixelREPA retains the JiT-H inference architecture because its masked adapter is used only during training \citep{shin2026pixelrepa}.

The RAE DiT-XL and DiT$^{\mathrm{DH}}$-XL counts are 238.12 and 323.16 GFLOPs, computed from the official 256-token architectures, including spatially conditioned modulation in the wide head. HyperDiT-XL's 551.4 GFLOPs are an estimate from the local implementation of its published configuration, including both branches, connectors and registers. RAEv2 uses 875.22M parameters and 475.00 GFLOPs, counted from its released DINOv3-L $K=7$ configuration, including the internal guidance head and excluding the representation encoder and decoder. AsymFlow uses its reported 363 GFLOPs. Parameter counts in the table otherwise follow the cited papers. These counts exclude pointwise nonlinearities and normalization, and describe model computation rather than hardware runtime.

\paragraph{Scaling-model costs.}
For SiHC, the counts include 32 context tokens from the block specified in Table~\ref{tab:sihc-model-config} and independent read/write maps at both sublayers. The DiT-only GFLOPs in Table~\ref{tab:sihc-scaling} are analytical estimates using DiP's published depths and widths, 256 tokens, the released SwiGLU backbone, and direct $16\times16$ input/output projections. We count attention products, feed-forward projections and AdaLN modulation under the same convention as SiHC. DiP's parameter counts, FID and IS are taken directly from its Table 2.

\paragraph{Controlled-model costs.}
For the main progression in Table~\ref{tab:progression}, we count model parameters, including frozen positional parameters where present, and add leading residual-interface arithmetic to the matrix-operation costs. The plain DiT uses 131.69M parameters and 46.60 GFLOPs. Copied-input and spatial mHC use 133.46M / 133.16M parameters, both at 47.73 GFLOPs. SiHC uses 24 sublayer read/write pairs: $8\times8$ / $4\times4$ patches give 130,754,496 / 130,975,536 parameters and 46.671 / 46.897 GFLOPs. Adding 32 in-context class tokens from the fifth block gives 131,000,112 parameters and 50.949 GFLOPs: their positional embeddings add 24,576 parameters, and attention/MLP computation on the longer sequence adds 4.052 GFLOPs. The static scalar-map estimate holds the $8\times8$ SiHC backbone fixed and replaces each pair of $C\times S$ feature-wise maps with two $S$-entry scalar maps. This gives 130,607,232 parameters and the same 46.671 GFLOPs. Feature-wise access adds 147,264 parameters while operating on the same residual-state coordinates.

\subsection{Linear-interface arithmetic cost}
\label{app:interface-cost}

A feature-wise read/write pair uses $2CS$ parameters and $4MCS$ leading FLOPs, counting a multiply-add as two FLOPs. A depth-$L$ sublayerwise model has $2L$ pairs, totaling $4LCS$ parameters and $8LMCS$ FLOPs. With $L=12$, $C=768$, $S=16$ and $M=256$, these are $0.590$M parameters and $0.302$G FLOPs. Scalar maps share coefficients across channels, reducing the interface parameter count to $4LS$ while retaining the same leading arithmetic over $MSC$ activations. The earlier blockwise variant has half as many read/write pairs (Appendix~\ref{app:granularity}).
These counts use the unfused recurrence to compare architectures under a common convention; stage fusion can change the executed arithmetic (Appendix~\ref{app:fusion}). For mHC, matrix-operation counting already includes the read, so the interface correction adds only the previously uncounted write contribution. Normalization and pointwise nonlinearities are excluded throughout.

\section{Read/Write Interface and Residual Paths}
\label{app:grad}
\begin{figure}[!htbp]
\centering
\resizebox{\linewidth}{!}{%
\begin{tikzpicture}[>=Latex,every node/.style={font=\scriptsize},
state/.style={draw,rounded corners,fill=blue!9,minimum height=7mm,minimum width=13mm,align=center},
work/.style={draw,rounded corners,fill=orange!12,minimum height=7mm,minimum width=19mm,align=center},
map/.style={draw,rounded corners,fill=green!8,minimum height=7mm,minimum width=13mm,align=center}]
\node[anchor=west,font=\small\bfseries] at (0,0.65) {(a) Blockwise: one read/write pair per block};
\node[state] (h) at (0.7,0) {$\mH_\ell$};
\node[map] (r) at (2.3,0) {Read $\mathcal R_\ell$};
\node[work] (a) at (4.2,0) {Attention\\$\vec a_\ell$};
\node[work] (m) at (6.65,0) {MLP\\input $\vec z_\ell+\vec a_\ell$};
\node[map] (w) at (9.2,0) {Write $\mathcal P_\ell$\\$\vec a_\ell+\vec m_\ell$};
\node[state] (o) at (11.3,0) {$\mH_{\ell+1}$};
\draw[->] (h)--(r);\draw[->] (r)--node[above] {$\vec z_\ell$}(a);\draw[->] (a)--(m);\draw[->] (m)--(w);\draw[->] (w)--(o);
\draw[->] (h.south)--++(0,-0.5)-|node[pos=0.3,below] {identity carry}(o.south);
\node[anchor=west,font=\small\bfseries] at (0,-1.4) {(b) Sublayerwise: write attention, then read again for MLP};
\node[state] (sh) at (0.4,-2.1) {$\mH_\ell$};
\node[map] (sr) at (1.85,-2.1) {Read $\mathcal R_\ell^{\mathrm A}$};
\node[work,minimum width=15mm] (sa) at (3.4,-2.1) {Attention};
\node[map] (sw) at (4.95,-2.1) {Write $\mathcal P_\ell^{\mathrm A}$};
\node[state] (smid) at (6.45,-2.1) {$\mH_{\ell+\frac12}$};
\node[map] (smr) at (7.95,-2.1) {Read $\mathcal R_\ell^{\mathrm M}$};
\node[work,minimum width=13mm] (smlp) at (9.45,-2.1) {MLP};
\node[map] (smw) at (10.95,-2.1) {Write $\mathcal P_\ell^{\mathrm M}$};
\node[state] (so) at (12.45,-2.1) {$\mH_{\ell+1}$};
\draw[->] (sh)--(sr);\draw[->] (sr)--(sa);\draw[->] (sa)--(sw);\draw[->] (sw)--(smid);\draw[->] (smid)--(smr);\draw[->] (smr)--(smlp);\draw[->] (smlp)--(smw);\draw[->] (smw)--(so);
\draw[->] (sh.south)--++(0,-0.5)-|(smid.south);\draw[->] (smid.south)--++(0,-0.8)-|(so.south);
\end{tikzpicture}}
\caption{\textbf{Two execution granularities of SiHC.} Blue boxes are persistent spatial states; orange boxes operate on the compact workspace. Blockwise SiHC writes the combined attention/MLP update once. Sublayerwise SiHC writes after attention and reads again for the MLP. Both use identity carry and the same spatial input/output assignment.}
\label{fig:sihc-granularity}
\end{figure}
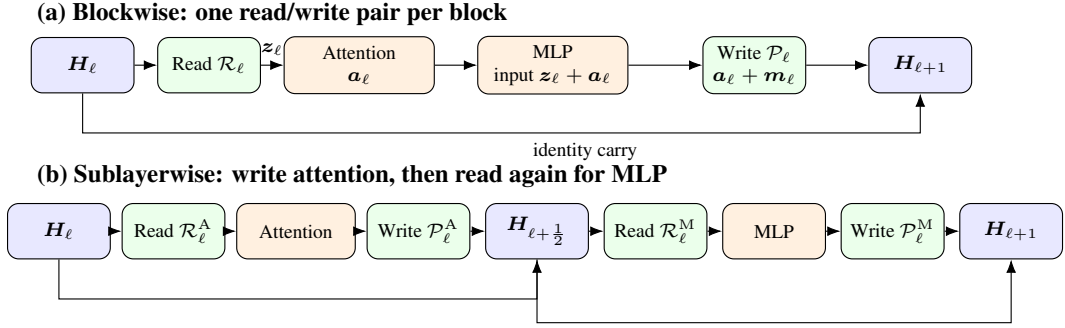

SiHC reads from and writes to its spatial states around each attention and MLP sublayer. We also specify the earlier blockwise form used for the embedding comparison in Table~\ref{tab:ablation-settings} (Figure~\ref{fig:sihc-granularity}).

\paragraph{Feature-wise read and write.}
\label{app:interface}
With elementwise multiplication $\odot$, Equation~\eqref{eq:sihc-read} becomes
\begin{align}
\vec h_\ell^{\mathrm{in}}&=\sum_s\Hpre^\ell[:,s]\odot\vec h_\ell^{(s)},\\
\vec h_{\ell+1}^{(s)}&=\vec h_\ell^{(s)}+\Hpost^\ell[:,s]\odot\Delta\vec h_\ell.
\end{align}
Each $C\times S$ map assigns a separate coefficient to every feature and stream.

\subsection{Sublayerwise updates and the earlier blockwise form}
\label{app:granularity}
Let $\mathcal R_\ell$ and $\mathcal P_\ell$ denote the feature-wise read and write above. Write $\mathcal A_\ell$ and $\mathcal M_\ell$ for the gated attention and MLP updates, including their normalization and time/class conditioning.

\paragraph{Sublayerwise.}
Attention and MLP each have their own read/write pair:
\begin{align}
\vec z_\ell^{\mathrm A}&=\mathcal R_\ell^{\mathrm A}(\mH_\ell),&
\mH_{\ell+\frac12}&=\mH_\ell+\mathcal P_\ell^{\mathrm A}\!\left(\mathcal A_\ell(\vec z_\ell^{\mathrm A})\right),\\
\vec z_\ell^{\mathrm M}&=\mathcal R_\ell^{\mathrm M}(\mH_{\ell+\frac12}),&
\mH_{\ell+1}&=\mH_{\ell+\frac12}+\mathcal P_\ell^{\mathrm M}\!\left(\mathcal M_\ell(\vec z_\ell^{\mathrm M})\right).
\end{align}
The MLP reads the attention-updated states through its own map, giving $2L$ connection events for $L$ blocks.

\paragraph{Earlier blockwise form.}
One read/write pair surrounds the combined attention and MLP update:
\begin{align}
\vec z_\ell&=\mathcal R_\ell(\mH_\ell),&
\vec a_\ell&=\mathcal A_\ell(\vec z_\ell),\\
\vec m_\ell&=\mathcal M_\ell(\vec z_\ell+\vec a_\ell),&
\mH_{\ell+1}&=\mH_\ell+\mathcal P_\ell(\vec a_\ell+\vec m_\ell).
\end{align}

\subsection{Gradients through one connection event}
The following derivatives apply to each connection event, with $\ell$ indexing events and $m=1,\ldots,M$ indexing computational cells. For blockwise SiHC, $\mathcal F_\ell$ combines attention and MLP updates, whereas for sublayerwise SiHC it denotes one of these updates.

\paragraph{Forward and backward paths.}
Let $g_\ell^{(s)}(m,c)=\partial\mathcal L/\partial h_{\ell,c}^{(s)}(m)$ be a scalar loss derivative. The gradient entering the nonlinear update and the gradient returned to its input satisfy
\begin{equation}
q_\ell(m,c)=\sum_s\Hpost^\ell[c,s]g_{\ell+1}^{(s)}(m,c),\qquad
\vec r_\ell=\mJ_\ell^\top\vec q_\ell.
\end{equation}
The vectors $\vec q_\ell,\vec r_\ell\in\mathbb R^{MC}$ stack all cells and features, and $\mJ_\ell\in\mathbb R^{MC\times MC}$ is the update Jacobian from its stacked read vectors to its stacked updates. This includes attention's interactions across cells. Identity carry then gives
\begin{equation}
g_\ell^{(s)}(m,c)=g_{\ell+1}^{(s)}(m,c)+\Hpre^\ell[c,s]r_\ell(m,c).
\end{equation}
A plain residual connection already has an identity path. SiHC widens the state preserved by this path while passing its nonlinear interaction through the smaller workspace.

\section{Kernel Fusion by Exact Stage-Wise Reordering}
\label{app:fusion}
\subsection{From repeated wide-state updates to a compact recurrence}
A literal SiHC update reads an $S$-stream state, computes a width-$C$ update, and writes the result back to every stream. The Transformer uses a compact workspace, but materializing each intermediate state repeatedly moves the wider $SC$ representation. Static feature-wise maps and identity carry let us collect these writes over a stage and evaluate the same recurrence through the compact updates.

For either granularity in Appendix~\ref{app:granularity}, let a stage contain $T$ residual updates, indexed by $\ell=0,\ldots,T-1$. Flatten batch and computational-cell indices into $q=1,\ldots,Q$, where $Q=B_{\mathrm{batch}}M$. Write $X_{\ell,qsc}$ for the retained state, $A_{\ell cs}=\Hpre^\ell[c,s]$ and $P_{\ell cs}=\Hpost^\ell[c,s]$. The literal operations are
\begin{equation}
Z_{\ell qc}=\sum_s A_{\ell cs}X_{\ell qsc},\qquad
\Delta Z_\ell=\mathcal F_\ell(Z_\ell,t),\qquad
X_{\ell+1,qsc}=X_{\ell qsc}+P_{\ell cs}\Delta Z_{\ell qc}.
\end{equation}
Unrolling the identity carry gives $X_{\ell qsc}=X_{0qsc}+\sum_{j<\ell}P_{jcs}\Delta Z_{jqc}$. Substitute this into the read and define
\begin{equation}
B_{\ell qc}=\sum_s A_{\ell cs}X_{0qsc},\qquad
\Gamma_{\ell jc}=\sum_s A_{\ell cs}P_{jcs}\quad(j<\ell).
\end{equation}
Then the forward pass is
\begin{align}
Z_{\ell qc}&=B_{\ell qc}+\sum_{j<\ell}\Gamma_{\ell jc}\Delta Z_{jqc},
&\Delta Z_\ell&=\mathcal F_\ell(Z_\ell,t),\label{eq:fused-read}\\
X_{\mathrm{out},qsc}&=X_{0qsc}+\sum_{\ell=0}^{T-1}P_{\ell cs}\Delta Z_{\ell qc}.
\end{align}
The first equation computes each compact workspace in sequence, and the second reconstructs all spatial states at the stage boundary (Figure~\ref{fig:fusion}). Since each workspace is the same algebraic function of preceding updates, attention and MLP computation receive the same inputs as in literal execution. Floating-point summation follows the fused reduction order.

This is \emph{late materialization}: each spatial state remains logically defined by its stage input and accumulated writes, while subsequent reads are evaluated directly from that representation. The expanded residual-stream bandwidth is therefore available without storing a separate wide tensor at every update. At the stage boundary, the final kernel accumulates each output tile on chip and writes it back once, avoiding the repeated wide-state transfers of literal execution.

\begin{figure}[!htbp]
\centering
\begin{tikzpicture}[>=Latex,every node/.style={font=\scriptsize},
wide/.style={draw,rounded corners,fill=blue!8,align=center,minimum width=16mm,minimum height=9mm},
compact/.style={draw,rounded corners,fill=orange!12,align=center,minimum width=24mm,minimum height=9mm},
op/.style={draw,rounded corners,fill=green!8,align=center,minimum width=24mm,minimum height=9mm}]
\node[anchor=west,font=\small\bfseries] at (0,0.65) {Literal execution: repeat for every update};
\node[wide] (x) at (0.8,0) {$X_\ell$\\$Q\times S\times C$};
\node[op] (read) at (3.0,0) {read $A_\ell$\\reduce over $S$};
\node[compact] (update) at (5.8,0) {$Z_\ell\to\mathcal F_\ell\to\Delta Z_\ell$\\$Q\times C$};
\node[op] (write) at (8.7,0) {write $P_\ell$\\add identity carry};
\node[wide] (next) at (11.1,0) {$X_{\ell+1}$\\$Q\times S\times C$};
\draw[->] (x)--(read);\draw[->] (read)--(update);\draw[->] (update)--(write);\draw[->] (write)--(next);
\node[anchor=west,font=\small\bfseries] at (0,-1.1) {Fused stage: retain compact updates, reconstruct once};
\node[wide] (base) at (0.8,-2.05) {$X_0$\\$Q\times S\times C$};
\node[op] (bases) at (3.0,-2.05) {1. All base reads\\$B_0,\ldots,B_{T-1}$};
\node[compact,minimum width=34mm,minimum height=13mm] (rec) at (6.1,-2.05) {2. For $\ell=0,\ldots,T-1$\\$Z_\ell=B_\ell+\sum_{j<\ell}\Gamma_{\ell j}\odot\Delta Z_j$\\$\Delta Z_\ell=\mathcal F_\ell(Z_\ell,t)$};
\node[op] (fin) at (9.6,-2.05) {3. Final writes\\$X_0+\sum_\ell P_\ell\odot\Delta Z_\ell$};
\node[wide] (out) at (12.25,-2.05) {$X_{\mathrm{out}}$};
\draw[->] (base)--(bases);\draw[->] (bases)--(rec);\draw[->] (rec)--(fin);\draw[->] (fin)--(out);
\draw[->] (base.south)--++(0,-0.65)-|(fin.south);
\node[font=\scriptsize,fill=white] at (5.7,-2.95) {stage identity carry};
\end{tikzpicture}
\caption{\textbf{Fusion removes intermediate wide-state materialization.} Blue tensors hold the $S$ persistent streams; orange tensors have workspace width $C$. All base reads are computed from the same stage input. The triangular recurrence restores the effect of earlier writes before each nonlinear update. Only the final operation reconstructs the wide state.}
\label{fig:fusion}
\end{figure}

\subsection{Three operation families and stage boundaries}
Table~\ref{tab:fusion-ops} gives the tensors and launch responsibilities. With $4\times4$ subpatches, the ImageNet state layout is $B_{\mathrm{batch}}\times64\times64\times C$, regrouped into $M=256$ cells with $S=16$ states each. Read and write coefficients are shared across cells and samples. They initialize to $A_{\ell cs}=1/S$ and $P_{\ell cs}=1$ and are learned independently at every update.

\begin{table}[!htbp]
\centering\small
\caption{\textbf{Fused interface operations.} $T$ counts residual updates in one stage and $Q=B_{\mathrm{batch}}M$.}
\label{tab:fusion-ops}
\begin{tabularx}{\linewidth}{@{}lXl@{}}
\toprule
Operation & Computation / implementation & Output shape \\
\midrule
Coefficient contraction & Reduce $A_\ell P_j$ over streams for $j<\ell$ & $T\times T\times C$ \\
All base reads & Read $X_0$ with every $A_\ell$; separate outputs feed compiled consumers & $T$ tensors $Q\times C$ \\
Active triangular sum & At update $\ell$, accumulate the previous $\ell$ updates with $\Gamma_{\ell j}$ & $Q\times C$ \\
Nonlinear update & Attention/MLP, normalization and conditioning on the compact workspace & $Q\times C$ \\
Final accumulation & Reduce all writes, add $X_0$, store the final spatial states & $Q\times S\times C$ \\
\bottomrule
\end{tabularx}
\end{table}

The CUDA fast path implements the base reads, active triangular sums and final accumulation as custom Triton operations. A distinct active-prefix operation fixes the number of preceding updates at compile time. The final kernel accumulates all writes in registers before storing each output element. Attention, RMSNorm, AdaLN and SwiGLU remain standard PyTorch operations captured by \texttt{torch.compile}. Thus one stage contains multiple launches and sequential nonlinear updates.

The blockwise kernels specialize to 8- and 12-block stages. The sublayerwise variant has independent attention and MLP connections and uses 16- and 24-event kernels for the same stage lengths. Its $T$ therefore counts twice as many connection events. At a stage boundary, $X_{\mathrm{out}}$ becomes the next $X_0$, resetting the update history. In-context class tokens are introduced at their configured start block. Stage sizes, connection granularity and context insertion are part of the model configuration.

\paragraph{Numerical accumulation.}
Forward reductions accumulate in FP32 and store results in the workspace dtype. Read/write coefficient gradients also accumulate in FP32 across samples and spatial positions.

\subsection{Backward propagation through the fused recurrence}
Gradients pass through the base reads, the triangular recurrence and the final write. For a loss applied to the stage output, let $\bar X_{qsc}=\partial\mathcal L/\partial X_{\mathrm{out},qsc}$ and let $G_{\ell qc}=\partial\mathcal L/\partial Z_{\ell qc}$ denote the total read gradient after reversing all later updates. Reverse execution visits $\ell=T-1,\ldots,0$. The gradient of the compact update is
\begin{equation}
D_{\ell qc}=\sum_s P_{\ell cs}\bar X_{qsc}
+\sum_{k>\ell}\Gamma_{k\ell c}G_{kqc},\qquad
G_\ell=\mJ_\ell^\top D_\ell,
\end{equation}
where $\mJ_\ell$ is the full nonlinear-update Jacobian, including attention across cells. The interface gradients are
\begin{align}
\bar\Gamma_{\ell jc}&=\sum_qG_{\ell qc}\Delta Z_{jqc},\quad j<\ell,\\
\frac{\partial\mathcal L}{\partial X_{0qsc}}
&=\bar X_{qsc}+\sum_\ell A_{\ell cs}G_{\ell qc},\\
\frac{\partial\mathcal L}{\partial A_{\ell cs}}
&=\sum_q G_{\ell qc}X_{0qsc}+\sum_{j<\ell}\bar\Gamma_{\ell jc}P_{jcs},\\
\frac{\partial\mathcal L}{\partial P_{jcs}}
&=\sum_q\bar X_{qsc}\Delta Z_{jqc}+\sum_{\ell>j}\bar\Gamma_{\ell jc}A_{\ell cs}.
\end{align}
These expressions identify the reductions required by the custom backward kernels. In particular, $\Gamma$ remains connected to both learned maps, so its gradients train the same read/write coefficients as literal execution. Checkpointing nonlinear activations can be combined with this interface schedule.

\subsection{Storage and arithmetic}
Literal autograd may retain $T$ wide states for its read-weight gradients, costing $TQSC$ elements. The reordered interface stores the stage input and width-$C$ base reads/updates, giving $O(QSC+TQC+T^2C)$ interface storage. Nonlinear attention/MLP activations are additional to both.

Fusion reduces intermediate state materialization and reorganizes memory access and launches. The leading interface work, counting a multiply-add as two FLOPs, is approximately
\begin{equation}
4TQCS+T(T-1)QC+T(T-1)CS.
\end{equation}
The first term computes base reads and final writes. The second is the triangular accumulation, and the third constructs its coefficients. Literal read/write execution has leading cost $4TQCS$. Short stages bound the additional triangular work. In the fused final accumulation, the wide input is read and the output is stored once; base-read kernels also access the stage input.

\subsection{Measured runtime and memory}\label{app:fusion-benchmark}
Table~\ref{tab:fusion-matched} compares literal PyTorch routing with fused routing under \texttt{torch.compile}, holding topology, size and recomputation fixed. Models use $256\times256$ inputs, $4\times4$ subpatches, a $16\times16$ workspace grid and 32 in-context tokens. B/L/XL/H have 12/24/40/32 blocks of width 768/1024/1024/1280. Both implementations use no recomputation for B/L and MLP recomputation for XL/H.

Measurements use one H100 80GB per trial, batch size 128, FP32 parameters, BF16 autocast and Flash SDPA, with PyTorch 2.9.1+cu128 and Triton 3.5.1. Training measures forward, loss, backward, fused AdamW and two FP32 EMA updates under default compilation and one-rank DDP. Inference measures one denoiser forward under \texttt{reduce-overhead} compilation. Each entry is the median of three fresh-process trials with eight warmup and 50 measured steps, excluding compilation, data loading, checkpoint I/O and REPA. Memory is peak allocated memory after training warmup, or including CUDA Graph setup for inference.

\begin{table}[!htbp]
\centering\small\setlength{\tabcolsep}{6pt}
\caption{\textbf{Compiled SiHC with and without fused routing.} Batch size 128 on H100 80GB. Training measures an optimizer step with two EMAs, and inference measures one denoiser forward. Memory is peak allocated GiB. OOM denotes failure at the same requested batch size.}
\label{tab:fusion-matched}
\begin{tabular}{@{}llrrrr@{}}
\toprule
& & \multicolumn{2}{c}{Training} & \multicolumn{2}{c}{Inference}\\
\cmidrule(lr){3-4}\cmidrule(l){5-6}
Size & Routing & ms & GiB & ms & GiB\\
\midrule
\multicolumn{6}{@{}l}{\textit{Blockwise}}\\
B & Torch & 123.42 & 26.82 & 35.90 & 3.05\\
B & Fused & 97.05 & 18.90 & 26.54 & 3.01\\
L & Torch & 365.24 & 70.83 & 102.17 & 5.05\\
L & Fused & 292.77 & 50.33 & 76.76 & 4.99\\
XL & Torch & OOM & OOM & 171.47 & 6.19\\
XL & Fused & 527.67 & 59.04 & 129.73 & 6.13\\
H & Torch & OOM & OOM & 191.16 & 7.71\\
H & Fused & 602.97 & 63.06 & 149.61 & 7.60\\
\midrule
\multicolumn{6}{@{}l}{\textit{Sublayerwise}}\\
B & Torch & 210.38 & 30.18 & 52.27 & 3.68\\
B & Fused & 113.73 & 19.82 & 28.82 & 3.25\\
L & Torch & OOM & OOM & 146.50 & 5.93\\
L & Fused & 330.64 & 52.07 & 83.48 & 5.44\\
XL & Torch & OOM & OOM & 247.14 & 7.07\\
XL & Fused & 573.15 & 62.07 & 137.36 & 6.08\\
H & Torch & OOM & OOM & 266.41 & 8.86\\
H & Fused & 682.42 & 67.18 & 157.72 & 8.19\\
\bottomrule
\end{tabular}
\end{table}

For sublayerwise B, fusion gives a $1.85\times$ training-step speedup and reduces peak allocated memory from $30.18$ to $19.82$ GiB. Single-forward inference speedups range from $1.69\times$ to $1.81\times$ across the four sublayerwise sizes. The anonymous source package provides reference implementations and numerical tests for the fused forward and backward operations.

\section{Linear Probes of Next-Layer Inputs}
\label{app:probes}

The probes measure how readily an ImageNet class can be decoded from the representation supplied to the next attention sublayer. They use frozen epoch-200 raw backbone weights, with true labels supplied only to the linear classifiers. All backbones receive null class conditioning. There is no additional noise injected into hidden states.

\paragraph{Input and time conventions.}
We form $\vec z=(1-\alpha)\vec x+\alpha\vec\epsilon$ at noise coefficients $\alpha\in\{0.25,0.5,0.75\}$. Pixel inputs are RGB images in $[-1,1]$, and RAE inputs are normalized DINOv2-B latents. The paper's clean coefficient is $t=1-\alpha$. The pixel implementations receive this $t$, while the RAE implementation uses the opposite time convention and receives $\alpha$. Each noise level has independently trained probes. The main results evaluate the same noise level used during probe training.

\paragraph{Extraction and fitting.}
At each boundary after blocks 1--11, the plain DiT supplies its residual state, while SiHC supplies the next attention read workspace $\{\vec h_\ell^{\mathrm{in}}(m)\}_{m=1}^{M}$. All are recorded before AdaLN. We apply non-affine RMS normalization to each patch token, average the tokens spatially, and fit an independent affine $768\rightarrow1000$ classifier at every boundary. Boundary 11 precedes block 12 and is the fixed boundary used in Table~\ref{tab:residual-probes}. All its entries use matched input noise, and SiHC assigns states to $4\times4$ subpatches.

\begin{table}[!htbp]
\centering\small
\caption{\textbf{Linear-probe fitting and evaluation.} Every boundary and noise level has an independent affine classifier.}
\label{tab:probe-protocol}
\begin{tabularx}{\linewidth}{@{}lX@{}}
\toprule
Setting & Hidden-state probes \\
\midrule
Frozen backbone & Raw epoch 200; null-class conditioning \\
Data & All 1,281,167 training and 50,000 validation images \\
Image preprocessing & ADM center crop; no random horizontal flips \\
Noise coefficient & $\alpha=0.25,0.50,0.75$, with paper time $t=1-\alpha$ \\
Feature / head & Per-token RMS ($10^{-6}$) then GAP; affine $768\to1000$ \\
Training & 40 epochs; batch size 4,096 \\
Optimizer & AdamW, default betas $(0.9,0.999)$, zero weight decay \\
Learning rate & Cosine from 0.01 to 0.0001 \\
Training noise & Fresh Gaussian noise each epoch; paired shuffle and noise across models \\
Random seeds & Training base 700001; validation 800001, 800002, 800003 \\
Precision & BF16 backbone; FP32 normalization, pooling and classifiers \\
Evaluation & Matched training/validation noise coefficient; mean and SD across three noise draws \\
\bottomrule
\end{tabularx}
\end{table}

\begin{table}[!htbp]
\centering\small\setlength{\tabcolsep}{5pt}
\caption{\textbf{State organization and prediction space shape class readability.} Entries give Top-1 / Top-5 accuracy (\%) at the last measured boundary. Columns increase input noise, and larger values indicate more linearly readable class information.}
\label{tab:residual-probes}
\begin{tabular}{@{}lrrr@{}}
\toprule
Model & 25\% noise & 50\% noise & 75\% noise \\
\midrule
Pixel, $\vec x$ & 25.32 / 46.45 & 27.54 / 49.64 & 22.82 / 43.91 \\
Pixel, $\vec v$ & 5.38 / 14.51 & 4.56 / 12.68 & 3.47 / 10.17 \\
Pixel SiHC, $\vec v$ & 25.51 / 46.68 & 27.81 / 50.32 & 22.88 / 43.98 \\
DINO, $\vec x$ & 80.15 / 95.47 & 80.09 / 95.41 & 80.06 / 95.36 \\
DINO, $\vec v$ & 81.09 / 95.74 & 81.07 / 95.81 & 80.93 / 95.77 \\
\bottomrule
\end{tabular}
\end{table}

\begin{figure}[!htbp]
\centering\includegraphics[width=\linewidth]{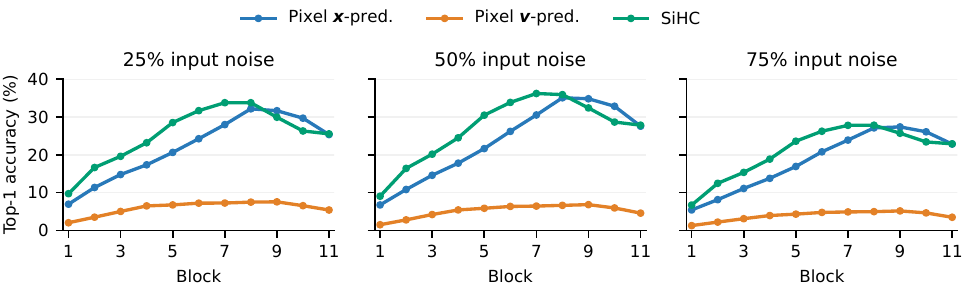}
\par\vspace{-5pt}
\caption{\textbf{SiHC recovers clean-like class readability while predicting velocity.} Panels fix the input-noise level. Linear-probe Top-1 accuracy measures class information through depth: SiHC approaches pixel $\vec x$-prediction and remains well above pixel $\vec v$-prediction. Shading denotes standard deviation across noise draws.}
\label{fig:sihc-probes}
\end{figure}

At $50\%$ input noise, the depth maxima of SiHC, pixel $\vec x$-prediction and pixel $\vec v$-prediction in Figure~\ref{fig:sihc-probes} are $36.18\%$, $35.05\%$, and $6.80\%$, respectively.

\begin{figure}[!htbp]
\centering\includegraphics[width=\linewidth]{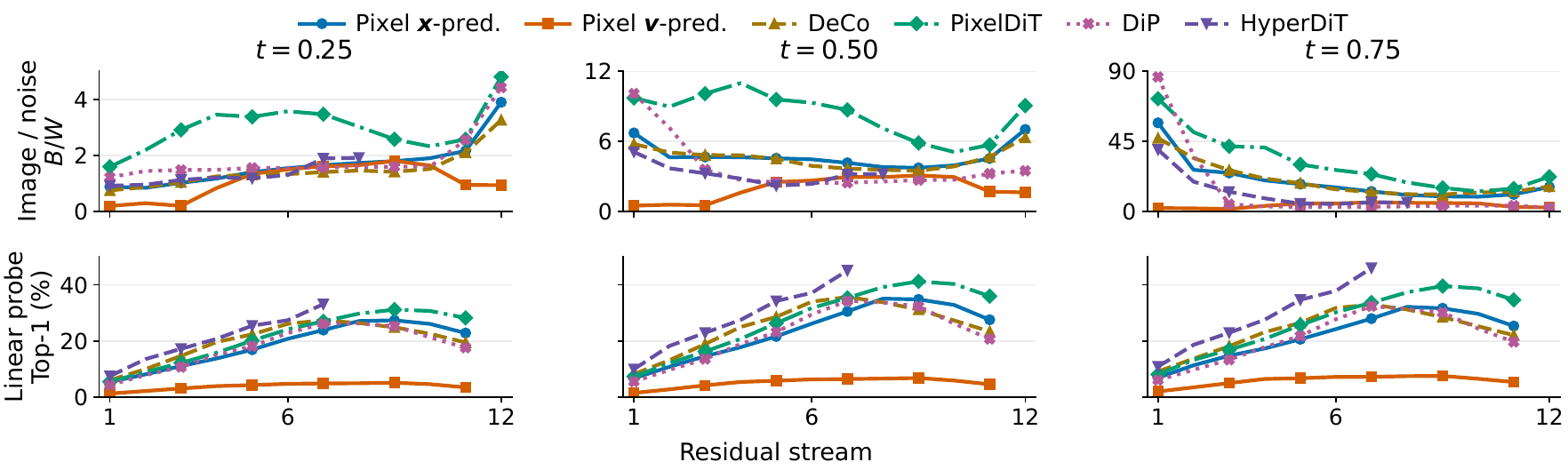}
\par\vspace{-5pt}
\caption{\textbf{Separate pixel paths support stronger semantic representations.} Columns fix the clean coefficient $t$. Top: between-image variation relative to noise ($B/W$). Bottom: linear-probe Top-1 accuracy through depth. Compare the decoupled $\vec v$-prediction models with pixel $\vec x$- and $\vec v$-prediction. Shading denotes standard deviation across noise draws.}
\label{fig:normalized-sensitivity}\label{fig:representation-probes}
\end{figure}

\paragraph{Whitened and decoupled models.}
Whitened models are probed in their own training space with the same protocol in Table~\ref{tab:probe-protocol}. Decoupled models expose the semantic state before the next block. The plain DiT, SiHC, DeCo, PixelDiT and DiP have 11 measured boundaries; the eight-block HyperDiT has seven. All features are recorded before the next attention normalization and modulation. Figure~\ref{fig:representation-probes} pairs the decoupled probes with their $B/W$ ratios.

\begin{figure}[!htbp]
\begin{minipage}[c]{0.36\linewidth}
\centering\includegraphics[width=\linewidth]{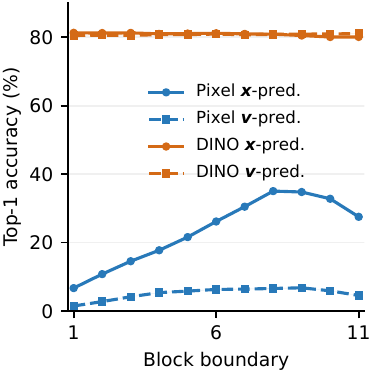}
\end{minipage}\hfill
\begin{minipage}[c]{0.60\linewidth}
\caption{\textbf{DINO-space layer inputs retain high class readability under both prediction targets.} Linear-probe accuracy follows block boundaries for DINO (orange) and pixel (blue) features. Solid lines use $\vec x$-prediction and dashed lines use $\vec v$-prediction. All curves use $50\%$ input noise. Shading denotes standard deviation across noise draws. The input control is reported in Table~\ref{tab:dino-input-probe}.}
\label{fig:rae-pixel-probes}
\end{minipage}
\end{figure}

\paragraph{Class readability in DINO space.}
Both plain DINO models have final-boundary probe accuracies above $80\%$ at all three measured noise levels (Table~\ref{tab:residual-probes}). Figure~\ref{fig:rae-pixel-probes} compares their depth profiles with pixel-space models, using the same extraction and fitting protocol.

At $50\%$ input noise, the final measured probe of DINO $\vec v$-prediction reaches $81.07\%$ Top-1 accuracy, compared with $4.56\%$ for pixel $\vec v$-prediction. The DINO input control reaches $80.49\%$ before the denoiser (Tables~\ref{tab:residual-probes} and~\ref{tab:dino-input-probe}).

\paragraph{DINO input control.}
The input control uses the official RAE encoder with DINOv2-with-registers base. Encoding resizes to 224, discards the class and register tokens, applies the encoder's non-affine final LayerNorm and the RAE latent normalization, and returns $256\times768$ features. The classifier acts on their spatial mean. Corruption is applied to the encoded latent features.

\begin{table}[H]
\centering\small
\caption{\textbf{Noisy DINO inputs retain strong class information.} Rows increase input noise. Top-1 and Top-5 measure linear-probe accuracy, and $\pm$ denotes standard deviation across noise draws.}
\label{tab:dino-input-probe}
\begin{tabular}{@{}lrr@{}}
\toprule
Matched input noise & Top-1 (\%) & Top-5 (\%) \\
\midrule
0\% & 79.372 & 95.036 \\
25\% & $79.872\pm.037$ & $95.267\pm.011$ \\
50\% & $80.485\pm.072$ & $95.545\pm.011$ \\
75\% & $80.820\pm.088$ & $95.847\pm.042$ \\
\bottomrule
\end{tabular}
\end{table}

The input-control probes use seed 42 and four zero-initialized affine classifiers, one for each noise level including zero noise. They train for 40 epochs with batch 4,096, AdamW betas $(0.9,0.999)$, $\epsilon=10^{-8}$, zero weight decay and cosine learning rate from 0.01 to 0.0001. Computation is FP32 with TF32 disabled. Evaluation uses the fixed epoch-40 checkpoint and three matched-noise draws on all 50,000 validation images. The input readout is direct GAP of normalized RAE latents.

\end{document}